\documentclass[journal]{IEEEtran}

\usepackage[utf8]{inputenc} % allow utf-8 input
\usepackage[T1]{fontenc}    % use 8-bit T1 fonts
\usepackage{hyperref}       % hyperlinks
\usepackage{url}            % simple URL typesetting
\usepackage{booktabs}       % professional-quality tables
\usepackage{amsfonts}       % blackboard math symbols
\usepackage{nicefrac}       % compact symbols for 1/2, etc.
\usepackage{microtype}      % microtypography
\usepackage{amsmath,amssymb,amsfonts}
\usepackage{graphicx}
\usepackage{textcomp}
\usepackage{threeparttable}
\usepackage{lipsum}
\usepackage{bbm}
\usepackage{xcolor}
\usepackage{bm}
\usepackage{tabularx}
\usepackage{array}
\usepackage{booktabs}
\usepackage{cases}
\usepackage[skip=1pt, belowskip=0pt]{caption}
\usepackage{subcaption} % for subfigures
\usepackage{amsthm}
\usepackage{arydshln}
\usepackage{enumitem}
\usepackage{mathtools}
\usepackage{amsmath}
\usepackage{comment}
\usepackage{enumitem}  
\usepackage{multirow}
\usepackage[normalem]{ulem} % for strikeout
\usepackage{float}
\usepackage{cite}
\def\boxit#1{%
  \smash{\fboxsep=0pt\llap{\rlap{\fbox{\strut\makebox[#1]{}}}~}}\ignorespaces
}

\makeatletter
\def\BibTeX{{\rm B\kern-.05em{\sc i\kern-.025em b}\kern-.08em
    T\kern-.1667em\lower.7ex\hbox{E}\kern-.125emX}}
\allowdisplaybreaks

\author{%
    Sanja~Karilanova, 
    Subhrakanti~Dey, 
    Ayça~Özçelikkale \\
    Department of Electrical Engineering, Uppsala University, Sweden \\
    \{Sanja.Karilanova, Subhrakanti.Dey, Ayca.Ozcelikkale\}@angstrom.uu.se\\
\thanks{S. Karilanova and A. Özçelikkale acknowledge the support of Center for Interdisciplinary Mathematics (CIM), and AI4Research, both in Uppsala University.}
\thanks{The computations were enabled by resources provided by the National Academic Infrastructure for Supercomputing in Sweden (NAISS), partially funded by the Swedish Research Council through grant agreement no. 2022-06725.}
}

\definecolor{myblue}{rgb}{0.3328, 0.3539, 0.7758}
\definecolor{myblue2}{rgb}{0.0328, 0.0539, 0.4758}
\definecolor{mygreen2}{rgb}{ 0.0328 0.4758 0.0539} 
\definecolor{mygreen3}{rgb}{ 0.0328 0.1758 0.0539} 
\definecolor{myred}{rgb}{0.4758, 0.0328, 0.0539}
\definecolor{myred2}{rgb}{0.75, 0.0328, 0.0539}
\definecolor{mypurple}{rgb}{0.5, 0.2, 0.7}
\definecolor{myblack}{rgb}{0.0, 0.0, 0.0}

\newcommand{\balert}[1]{{\color{myblack}{#1}}}

\newcommand{\stablereset}{{\it{Stable+Reset}}}
\newcommand{\stablenoreset}{{\it{Stable+NoReset}}}
\newcommand{\unstablereset}{{\it{Unstable+Reset}}}
\newcommand{\unstablenoreset}{{\it{Unstable+NoReset}}}
\newcommand{\stable}{{\it{Stable}}}
\newcommand{\unstable}{{\it{Unstable}}}
\newcommand{\reset}{{\it{Reset}}}
\newcommand{\noreset}{{\it{NoReset}}}

\newtheorem{rem}{\bf{Remark}}

\theoremstyle{remark}

\newenvironment{remark}
{\par\noindent \rem \begin{itshape}\noindent}
{\end{itshape} \vspace{3pt}}

\newcommand{\Thetabf}{\bf{\Theta}}
\newcommand{\ActFnc}{f}

\newcommand{\Umem}[1]{u[#1]}
\newcommand{\Uad}[1]{v[#1]}
\newcommand{\UmemNoArg}{u} % \Umem without argument
\newcommand{\UadNoArg}{v} % \Uad without argument
\newcommand{\Sil}[1]{s_{out}[#1]}
\newcommand{\Sl}[1]{\bm{s}_{out}[#1]}

\newcommand{\Silo}{s_{out}}

\newcommand{\Lpre}{L_{\text{pre}}}
\newcommand{\Lpost}{L_{\text{post}}}
\newcommand{\vbfRst}{\bm{v}_{\text{rst}}}

\newcommand{\Abf}{\bm{A}}
\newcommand{\Bbf}{\bm{B}}
\newcommand{\Rbf}{\bm{R}}
\newcommand{\Cbf}{\bm{C}}

\newcommand{\CbfBias}{\bm{c}_{bias}}
\newcommand{\Dbf}{\bm{D}}
\newcommand{\Lambdabf}{\bm{\Lambda}}

\newcommand{\Ibf}{\bm{I}}
\newcommand{\Inbf}{i}
\newcommand{\Wbf}{\bm{W}}

\newcommand{\ybf}{\bm{y}}
\newcommand{\ybfi}{y_k}
\newcommand{\vbf}{\bm{v}}
\newcommand{\vbfi}{v_k}

\newcommand{\nout}{n_{out}}
\newcommand{\nin}{n_{in}}

\newcommand{\Hneurons}{h}

\newcommand{\Nstate}{n}

\newcommand{\RstValue}{r_{\text{scale}}} % {r_{\text{rst}}} or simply {r}
\newcommand{\RstBias}{r_{\text{bias}}}
\newcommand{\RstFncC}{\mathcal{R}_{c}}
\newcommand{\RstFncA}{\mathcal{R}_{a}}
\newcommand{\bigO}{\mathcal{O}}
\newcommand{\cin}{c_{in}}
\newcommand{\cout}{c_{out}}

\newcommand{\R}{\mathbb{R}}
\newcommand{\C}{\mathbb{C}}

\mathchardef\myhyphen="2D

\begin{document}
\bstctlcite{IEEEexample:BSTcontrol}
\title{Low-Bit Data Processing Using Multiple-Output Spiking Neurons with Non-linear Reset Feedback}
\maketitle
\begin{abstract}
Neuromorphic computing is an emerging technology enabling low-latency and energy-efficient signal processing. A key algorithmic tool in neuromorphic computing is spiking neural networks (SNNs). SNNs are biologically inspired neural networks which utilize stateful neurons, and provide low-bit data processing by encoding and decoding information using spikes. Similar to SNNs, deep state-space models (SSMs) utilize stateful building blocks. However, deep SSMs, which recently achieved competitive performance in various temporal modeling tasks, are typically designed with high-precision activation functions and no reset mechanisms. To bridge the gains offered by SNNs and the recent deep SSM models, we propose a novel multiple-output spiking neuron model that combines a linear, general SSM state transition with a non-linear feedback mechanism through reset. Compared to the existing neuron models for SNNs, our proposed model clearly conceptualizes the differences between the spiking function, the reset condition and the reset action. The experimental results on various tasks, i.e., a  keyword spotting task, an event-based vision task and a sequential pattern recognition task, show that our proposed model achieves performance comparable to existing benchmarks in the SNN literature. Our results illustrate how the proposed reset mechanism can overcome instability and enable learning even when the linear part of neuron dynamics is unstable, allowing us to go beyond the strictly enforced stability of linear dynamics in recent deep SSM models.  
\end{abstract}

\begin{IEEEkeywords}
event-based signal processing, spiking neural networks (SNNs), state-space models (SSMs), reset-mechanism, neuromorphic.
\end{IEEEkeywords}

\section{Introduction}

Signal processing and machine learning methods are typically implemented in the conventional {\it{von Neumann}} computing architectures. The effectiveness of these methods often comes at the cost of high energy consumption for computation \cite{sudhakar_data_2023}.   
Additionally, several layers of data processing are desired in an increasing number of low-power internet of things (IoT) edge applications, such as large scale deployments for remote monitoring of industrial sites, and monitoring/surveillance with battery-powered drones. Neuromorphic computing, where insights from neuroscience are utilized for building computing architectures and processing data, has recently emerged as a promising solution for energy-efficient data processing 
\cite{Rajendran_2019, davies_advancing_2021, yik2024neurobench}. Accordingly, data processing on neuromorphic chips has shown to achieve significant gains in terms of low-latency and energy-efficiency compared to solutions implemented on the conventional chips for a range of applications, including constraint satisfaction problems and least absolute shrinkage and selection operator (LASSO) problems \cite{davies_advancing_2021}.

A key concept in neuromorphic computing is the encoding and decoding of information using spikes, i.e. impulses. The typical encoding scheme is binary spikes where in discrete-time data processing, the temporal information is represented as a sequence of $0$'s and $1$'s. In other words, the information is quantized with two levels, i.e. with 1-bit resolution.  Representation of information with 1-bit resolution allows conversion of multiplication operations to additions, resulting in lower number of arithmetic operations and power consumption \cite{yik2024neurobench}.

The typical computational model used in neuromorphic hardware is spiking neural networks (SNNs) \cite{davies_advancing_2021}, where the basic unit of computation is a spiking neuron.  In a standard artificial neural network, the input to a neuron is a continuous-valued variable and the neuron apply a static transformation to its input.
In contrast, the biological neurons have a state, such as a membrane voltage, and they process spike-trains inputs and produce spike trains outputs.
Spiking neurons in SNNs mimic these behavior. One of the most popular models is the Integrate and Fire (IF) model which integrates its input, and fires a spike and resets its state to its neutral value whenever it fires a spike 
\cite{Gerstner_Kistler_2002}.

Neuron models in SNNs are typically designed using biological plausibility based on neuroscience models \cite{Gerstner_Kistler_2002, baronig2024advancingspatiotemporalprocessingspiking, bittar2022surrogate, Fang_2021_ICCV}. Here, we combine insights from state space models (SSMs) in signal processing and machine learning, with insights from neuroscience to develop new neuron models to use in neural networks (NNs) and illustrate their performance. SSMs have been used in various successful signal processing methods, lately they have even obtained comparable performance to transformers in deep NN literature \cite{gu2022efficientlymodelinglongsequences,gu2022parameterization, smith2023simplified}. However, the typical SSMs in NNs do not use low-bit processing. In particular, in contrast to SNNs, where the output of the neurons are represented using spikes which are sparse in time and has limited amplitude resolution,  the outputs of SSMs in a typical deep SSM are continuous-valued. Motivated by the computational complexity and energy gains that can be obtained using low-bit data processing as well as promising energy efficiency results in neuromorphic computing \cite{Rajendran_2019, davies_advancing_2021, yik2024neurobench}, in this article we propose SSM based NN models with low-bit activation and hence communication between neurons.

A central question in our investigation is "How should we model low-bit SSM-based NNs?" Linear SSMs cannot model the nonlinear behavior existing in various signals, while non-linear SSMs are difficult to develop, interpret and use in data processing pipelines. Hence, we propose a SSM that combines a linear state-transition dynamics with a non-linear feedback through reset.

\subsection{Contributions}     
Our contributions can be summarized as follows:
\begin{itemize} 
    \item We propose a multiple-output stateful neuron model with a non-linear feedback mechanism through a state reset. In contrast to the traditional neuron models in SNN literature, this model allows a more general model structure through a general state transition matrix, multiple-outputs for the neuron and a general reset mechanism with learnable parameters. In contrast to typical SSM based models in NN literature, this neuron model brings low-bit information representation with spikes, non-linear feedback and allows instability in the SSM through the state transition matrix. 
    
    \item Our development generalizes the reset mechanism in SNN literature, which is typically based on neuroscience models and focuses on particular state variables.  Our development  clearly conceptualizes the differences between the spiking function, the reset condition,   the reset action. 
    
    \item \balert{Using the neuromorphic computing benchmark dataset \cite{yik2024neurobench} from the Multilingual Spoken Word Corpus (MSWC) dataset \cite{mswcdataset}, DVS-Gesture  dataset \cite{Amir_2017_CVPR} and sequential MNIST (sMNIST) dataset \cite{le2015simple}}, we illustrate that performance at the same level as the benchmark results in the literature can be obtained using the proposed  neuron model with non-linear feedback. 
    
    \item Our results illustrate that in order to have satisfactory performance, one does not need to have stable linear SSM dynamics, hence one does not need to constrain the eigenvalues of the  state transition matrix, which is a central concern in developing models in SSM-based NN literature. Our results provide a starting point for investigation of a richer family of SSM models in neural networks. 
\end{itemize}

We note that the typical neuron model in SNN literature is single output, with only a few recent exceptions \cite{Karilanova2025_NICE}. 
In contrast to the main-stream  single output neuron model in the SNN literature, here we consider a neuron model with multiple outputs. Different from the linear model in \cite{Karilanova2025_NICE} with no reset and stable linear dynamics, here we consider a non-linear feedback through a state reset mechanism with learnable parameters, and allow unstable state-transition matrices.   

\section{Related Work}
\label{sec:related-work}

The Integrate-and-Fire (IF) neuron, one of the most basic neuron models for SNNs, is closely related to the Time Encoding Machines \cite{lazar_perfect_2004}. 
Time Encoding Machines (TEMs) are event-based sensors \cite{lazar_perfect_2004}. Integrate-and-Fire (IF) TEM fires a spike when the integral of a moving window of a signal exceeds a certain threshold \cite{lazar_perfect_2004}. This IF behavior of TEMs is similar to IF behaviour in neurons, where the main difference is that TEMs are typically studied under continuous-amplitude inputs whereas IF neurons in SNNs typically use spike trains. Both IF TEMs and IF neurons  produce spike trains as output.  The operation of IF TEM can be directly associated with non-uniform sampling leading to sufficiency results for reconstruction of signals from TEM outputs \cite{lazar_perfect_2004}. Recently, 
researchers started to show growing interest in the TEM based event-based sampling framework, leading to a range of signal reconstruction results \cite{naaman_fri_tem_2021,thao_bandlimited_2022, adam_asynchrony_2022}.

State-space models (SSMs) are heavily utilized in the fields of signal processing and control for system modeling and modeling of temporal dependency. In deep learning, popular models for capturing long-term dependencies in temporal data have been  recurrent architectures, complex gated models and transformers among the few. In the last few years, SSM based deep neural networks have 
emerged as a promising simpler alternative by achieving competitive performance on a large set of long-range sequence modeling tasks, see S4 \cite{gu2022efficientlymodelinglongsequences} and S5 \cite{smith2023simplified}. The promise of such structures has also been shown under the neuromorphic low-bit resolution,  event-based data \cite{schöne2024scalableeventbyeventprocessingneuromorphic, soydan2024s7selectivesimplifiedstate}.

% Quantized SSMs
Recently, making SSM based NNs more computationally efficient has been studied using  quantization of the weights \cite{chiang2024quambaposttrainingquantizationrecipe}, and quantization of combination of weights, SSM parameters, activation function under both quantization-aware training and post-training quantization \cite{abreu2024qs5quantizedstatespace}, as well as under quantization aware fine-tuning \cite{meyer2024diagonalstructuredstatespace}.
% Connect quantized SSMs with SNNs
Quantization of the activation function, or the output of SSMs to 1-bit, is closely related to the inherent and biologically-inspired 1-bit information processing in SNNs. Hence, even more recently, there has been an effort to bridge SSMs and SNNs \cite{Karilanova2025_NICE,bal2024rethinking, stan2024learning, shen2024spikingssmslearninglongsequences} \cite{fabre2025structuredstatespacemodel}. However, drawing this bridge is not straightforward due to the linear nature of SSMs and recurrently non-linear nature of SNNs \cite{karilanova2024zeroshottemporalresolutiondomain}, especially due to the resetting of the membrane potential once a neuron has spiked. 
% Reset
To the best of our knowledge, no prior work has investigated modeling a reset in deep-SSM especially in a one-bit and multiple output neuron setting. 
% Unstability
In addition, the works in deep SSMs  have been restricted to stable initialization of SSMs in various ways \cite{gu2022parameterization} due to lack of mechanism that will stabilize the learning. In this paper we show that reset in deep SSMs can overcome unstable initialization and training.
\section{System Formulation}

\subsection{Preliminaries}
\label{sec:methods:preliminaries}

\subsubsection{General Spiking Neuron Model}
\label{sec:general_spiking_neuron}
A single general feed-forward discrete-time $\Nstate$-state dimensional spiking neuron can be defined as
\begin{subequations}
\label{eqn:general_spiking_neuron}
\begin{align}
\! \vbf[t+1] &=  \Abf \vbf[t] 
                - \Rbf \Sil{t}
                + \Bbf \Inbf[t]
                \label{eqn:general_spiking_neuron:line1}\\
\Sil{t} &= \ActFnc_{\Thetabf}(\vbf[t]) 
         =\begin{cases} 
          1  \text{ if }\vbf[t] \in \Thetabf \\
          0  \text{ otherwise}
          \end{cases}
                \label{eqn:general_spiking_neuron:line2}
\end{align}
\end{subequations}
where $\Inbf[t] \in \R^{1\times 1}$ is the input to the neuron, $\Silo[t] \in \R^{1\times 1}$ is the output of the neuron and $\vbf \in \R^{\Nstate \times 1}$ is the state variable of the neuron.
The matrices $\Abf \in \R^{\Nstate \times \Nstate}, \Bbf \in \R^{\Nstate \times 1}, \Rbf \in \R^{\Nstate \times 1}$ represents the leak, impact of the input and the  reset feedback, respectively.
The spiking behavior is described by $\ActFnc_{\Thetabf}(\cdot)$, where the neuron spikes when the state enters into the region described by $\Thetabf$. 
See Table \ref{tab:notation} for the key notation used in this paper.

The above formulation considers a type of soft-reset mechanism where the output spikes $\Sil{t}$ are multiplied by $\Rbf$ and subtracted from the state, see \eqref{eqn:general_spiking_neuron:line1}. Another reset mechanism that is popular in the SNN literature is hard-reset where the neuron state is set to a predefined reset state $\vbfRst$, e.g. to a state of all zeros, once spiking occurs \cite{davies_Loihi_2018}. Under such hard-reset, the $\Rbf \Sil{t}$ term is removed from \eqref{eqn:general_spiking_neuron}, and the below additional operation is performed: 
\begin{align}
\label{eqn:general_spiking_neuron_hard_reset}
\vbf[t+1] &=
         \begin{cases} 
          \vbfRst  \text{ if }\Sil{t} = 1 \\
          \vbf[t+1]  \text{ otherwise}
          \end{cases}
\end{align}

Spiking neural networks are built by connecting neurons such that the output spikes ${\bm{s}_{out}^L}[t]\in\R^{\Lpre\times1}$ from the previous (pre-synaptic) layer serve as inputs to the next (post-synaptic) layer $\Ibf[t]\in\R^{\Lpost \times 1}$, where $\Lpre$ and $\Lpost$ is the number of neurons in the pre and post synaptic layer. Specifically, $\Ibf[t]=\Wbf \times {\bm{s}_{out}^L}[t]$ where $\Wbf \in \R^{\Lpost \times \Lpre}$ is the weight matrix representing synaptic connection strength between the two layers.

%%%%%%%%%%%% Table - Notation %%%%%%%%%%%%
\begin{table}
\centering
\caption{Table of notation}
\label{tab:notation}
\newcolumntype{m}{>{\centering\arraybackslash}p{0.3 \linewidth}}
\begin{tabular}{mp{0.55\linewidth}}
 \hline
 Notation & Meaning \\
\hline
\multirow{2}{*}{$\Nstate$} & Dimension of a neuron \\
    & $=$ Number of variables in the state  \\
\hline
\multirow{2}{*}{$\nout$} & Output dimension  \\
    & $=$ Number of output channels of a neuron  \\

\hline
$  \vbf[t] \in R^{\Nstate}$       & State of the neuron \\
     \hline
$ \Inbf[t] \in R^{1}$       & Input of the neuron \\
       \hline
     $\Sl{t} \in R^{\nout}$       & Spike output of the neuron \\
   \hline
$\ActFnc(\cdot)$ & Activation/Spiking function \\
\hline
$\RstFncC(\cdot)$ & Reset condition parametrization \\
\hline
$\RstFncA(\cdot)$ & Reset action \\
\hline
$\cin$ & Number of input channels  for the network \\
\hline
$\cout$ & Number of output channels for the network \\
\hline
    $\Hneurons$ & Number of neurons in a hidden layer \\
\hline
\end{tabular}
\end{table}

\subsubsection{Popular Spiking Neuron Models}
\label{sec:popular_spiking_neuron}

% EXAMPLE - LIF NEURON
The discrete-time Leaky Integrate-and-Fire (LIF) neuron is defined as \cite{bittar2022surrogate,Gerstner_Kistler_2002}:
\begin{subequations}
\label{eqn:LIFneuron}
\begin{align}
      \label{eqn:LIF-transition}
\! \Umem{t+1} &= \alpha \Umem{t} - \alpha \theta \Sil{t}
                    + (1-\alpha) \Inbf[t] \\    \Sil{t} &= \ActFnc_\theta(\Umem{t})  % = \one(\Umem{t} \geq \theta)
         =\begin{cases} 
          1  \text{ if }\Umem{t} \geq \theta \\
          0  \text{ otherwise.}
          \end{cases}
         \label{eqn:LIF-spk}
\end{align}
\end{subequations}
where  $\Umem{t}$ is the state variable representing the membrane potential.
The LIF neuron is a special  case of the generalized spiking neuron in  \eqref{eqn:general_spiking_neuron}. The  LIF neuron is obtained by setting $\Nstate=1$, $\vbf=\UmemNoArg$, $\Abf=\alpha, \Bbf=(1-\alpha), \Rbf=\alpha \theta$, and ${\Thetabf}= \{\UmemNoArg \geq \theta\}$ where $\theta=1$ for a standard LIF neuron.

% EXAMPLE - adLIF NEURON
We now discuss the  Adaptive Leaky Integrate-and-Fire (adLIF) neuron, which is an extension of the LIF neuron with an additional  recovery state variable, $\Uad{t}$ \cite{bittar2022surrogate}. Discrete-time adLIF neuron is defined as follows
\begin{subequations}
\label{eqn:adLIFneuron}
\begin{alignat}{10}
\! \Umem{t+1} =& \alpha \Umem{t} - \alpha \theta \Sil{t}
                + (1-\alpha) \Inbf[t]  
                - (1-\alpha) \Uad{t}  \label{eqn:adLIFneuron_line1}\\ 
\! \Uad{t+1} =& a \Umem{t} + \beta \Uad{t}
                    + b \Sil{t} \label{eqn:adLIFneuron_line2}\\ 
\Sil{t} =& \ActFnc_\theta(\Umem{t})  
\label{eqn:adLIF-spk}
\end{alignat}
\end{subequations}
The adLIF neuron is a special  case of the generalized spiking neuron in  \eqref{eqn:general_spiking_neuron}. It is obtained by setting   $\Nstate=2$ and 
\begin{align} 
\vbf=\begin{bmatrix} \UmemNoArg \\ \UadNoArg \end{bmatrix},
\Abf=\begin{bmatrix}\alpha & -(1-\alpha)\\ a &\beta\end{bmatrix},
\Bbf=\begin{bmatrix} 1-\alpha \\ 0\end{bmatrix}, 
\Rbf=\begin{bmatrix} - \alpha \theta \\ b \end{bmatrix}. \notag
\end{align}

\begin{remark}
Although in Section~\ref{sec:general_spiking_neuron} we provide a general spiking neuron model with arbitrary number of $\Nstate$ state variables, we emphasize that the existing popular SNN models typically have a state dimension of $\Nstate=1$ (LIF) or $\Nstate=2$ (example adLIF) as illustrated in the above. Other examples of popular but less used spiking neuron models include Izikevich \cite{izhikevich2003simple} with $\Nstate=2$, and Hodgkin-Huxley \cite{Hodgkin1952} with $\Nstate=4$.
Recently, motivated by the developments in SSM models in neural networks,  a limited number of works \cite{bal2024rethinking, stan2024learning, Karilanova2025_NICE}, have investigated spiking neuron architectures with neurons with higher state dimensions. 
\end{remark}

\subsubsection{State Space Models}
\label{sec:ssm}

A discrete-time time-invariant linear SSM can be written as \cite{Gajic}:
\begin{subequations}
\label{eqn:lssm_generic}
\begin{align}
    \vbf[t+1] = \Abf_s \vbf[t] + \Bbf_s \Ibf[t] \label{eqn:lssm_generic_line1}\\ 
    \ybf[t] = \Cbf_s \vbf[t] + \Dbf_s \Ibf[t], \label{eqn:lssm_generic_line2}
\end{align} 
\end{subequations}
where  $\vbf[t], \Ibf[t], \ybf[t]$ denote  the state vector, input vector and output vector, respectively.  
Here, the state transition matrix $\Abf_s \in \R^{\Nstate \times \Nstate}$ describes the internal recurrent behavior of SSM. The  matrices $\Bbf_s \in \R^{\Nstate \times \nin}, \Cbf_s \in \R^{\nout\times \Nstate}, \Dbf_s \in \R^{\nout\times \nin}$ describe the interaction of the SSM through its inputs and outputs. By stacking layers that consist of a linear SSM,  i.e., \eqref{eqn:lssm_generic}, and element-wise nonlinearity applied to the output $\ybf$ of the linear SSM, SSM networks are formed \cite{gu2022efficientlymodelinglongsequences, gu2022parameterization, smith2023simplified}.

\subsection{Comparison of Spiking Neuron Models and Linear SSMs}
\label{sec:compare_SNN_SSM}

% SNNs vs SSMs 
We now compare the general spiking neuron model in Section \ref{sec:general_spiking_neuron} and linear SSM in Section \ref{sec:ssm}. 
% Similarities - Correspondences
The leak matrix $\Abf$ in \eqref{eqn:general_spiking_neuron:line1} and the state transition matrix $\Abf_s$ in \eqref{eqn:lssm_generic_line1} drive the linear part of the state transition in both models. Similarly, both $\Bbf$ and $\Bbf_s$  parameterize the impact of the input. 
% Differences - (i) nonlinear function + (ii) only based on membrane
However, one difference  is how the output is obtained.  In the spiking neuron the output is obtained by applying a non-linear spiking function on the raw state variable $\vbf[t]$, see  \eqref{eqn:general_spiking_neuron:line2}. In the linear SSM, the output is obtained by linear transformation of the state variable $\vbf[t]$, see \eqref{eqn:lssm_generic_line2}. Alternatively, in a deep SSM,  the output of the layer is obtained using a real, continuous-valued activation function on $\ybf[t]$ of \eqref{eqn:lssm_generic_line2} \cite{gu2022efficientlymodelinglongsequences, gu2022parameterization, smith2023simplified}.

% Differences - Reset
Another major difference of the models in Section \ref{sec:general_spiking_neuron} and  Section \ref{sec:ssm} is the lack of reset mechanism in the model of Section \ref{sec:ssm}. This is a key difference, which we focus on in this paper. The reset mechanism in the spiking neuron, as defined in \eqref{eqn:general_spiking_neuron:line1} or \eqref{eqn:general_spiking_neuron_hard_reset}, enforces state discontinuities, i.e. reset, as a response to spikes in its output. This type of non-linear system feedback is not present in the linear SSM model in \eqref{eqn:lssm_generic}. We note that  the element-wise non-linearity applied in deep SSM are only used to change the output of the layer and is not provided as a feedback to a layer; hence does not serve a reset mechanism.

\section{Proposed Multiple-output Neuron Model with Reset }

\subsection{Motivation}

We now discuss several aspects of our motivation behind the proposed neuron model.

\subsubsection{1-bit Spike-based Information Processing}
Neural networks operating under low-bit precision communication constraints must efficiently transmit information between neurons in the neural network. The spiking mechanisms in SNNs inherently adhere to this low-bit information processing paradigm, enabling energy-efficient information transfer. However, the performance of deep-SSM under such constraints remains under-explored, especially under the extreme 1-bit constraint for the communication between neurons.

\subsubsection{Neuroscience Perspective on Reset}
Biological neurons exhibit resetting dynamics after spiking, which plays a crucial role in  their encoding of temporal information.
A fundamental open question is how to formally incorporate reset mechanisms into SSM-based neural architectures and whether such mechanisms can enhance expressivity under 1-bit processing of temporal information.

\subsubsection{Beyond Enforced Stability}
SSM-based networks typically enforce a strictly stable dynamics both in the initialization as well as throughout training \cite{gu2022parameterization}. While this ensures well-behaved optimization in long-sequence operations, it limits the range of dynamical behaviors that can be represented. For example, strict stability conditions in the state transition may contribute to dead neuron problem; and fast decay of activity that has been initiated in the network due to specific temporal patterns, limiting the ability to capture long-range dependencies. In this paper, we allow unstable dynamics and investigate whether reset mechanisms can prevent divergence in such models, as well as the associated performance trade-offs.

\subsubsection{Beyond Linearity}
SSM-based neurons are typically modeled as linear systems. However, insights from signal processing, control theory, and neuroscience suggest that nonlinear dynamics could be fundamental to biological and engineered systems. Introducing reset mechanism to linear SSM-based neurons provides a natural extension toward nonlinearity, offering a stepping stone to more expressive and biologically plausible SSM-based architectures.

\subsection{Proposed Model}
\label{sec:proposed_model}

Our proposed multiple-output SSM-based neuron with general reset mechanism is defined as:
\begin{subequations}
\label{eqn:general_spiking_ssm_neuron:SIMO}
\begin{align}
\! 
\vbf[t+1] &=  \Abf \vbf[t] + \Bbf \Inbf[t] 
    \label{eqn:general_spiking_ssm_neuron:SIMO:state} \\
\ybf[t] &=\Cbf \vbf[t] + \CbfBias
    \label{eqn:general_spiking_ssm_neuron:SIMO:obsy} \\
\Sl{t} &= \ActFnc_\theta(\ybf[t]) 
    \label{eqn:general_spiking_ssm_neuron:SIMO:obsspike} \\
\vbf[t+1] &= \begin{cases} 
          \RstFncA (\vbf[t+1])  \text{ if } \RstFncC(\ybf[t]) \in \Thetabf \\
          \vbf[t+1]  \text{ otherwise}
          \end{cases}
    \label{eqn:general_spiking_ssm_neuron:SIMO:reset}  
\end{align}
\end{subequations}
% input/output
where $\Inbf[t] \in \R^{1 \times 1}$ is the input to the neuron at time step $t$ and $\Sil{t} \in \R^{\nout \times 1}$ is the spiking output of the neuron at time step $t$ produced by the spiking function $\ActFnc_\theta(\cdot)$. 
% SSM parameters
The internal linear dynamics is parametrized by the matrix parameters $\Abf \in \C^{\Nstate \times \Nstate}$, $\Bbf \in \R^{\Nstate \times 1}$ whereas  $\Cbf \in \C^{\nout \times \Nstate}$, and $\CbfBias  \in \C^{\nout \times 1}$ contributes to the output. 
Here, motivated by deep SSM literature, we consider a possibly complex-valued state dynamics through complex $\Abf$ whereas the input and output are real-valued through real-valued $\Bbf$ and $\ActFnc_\theta(\cdot)$, respectively.

\begin{figure}
    \centering
    \includegraphics[width=0.49\textwidth]{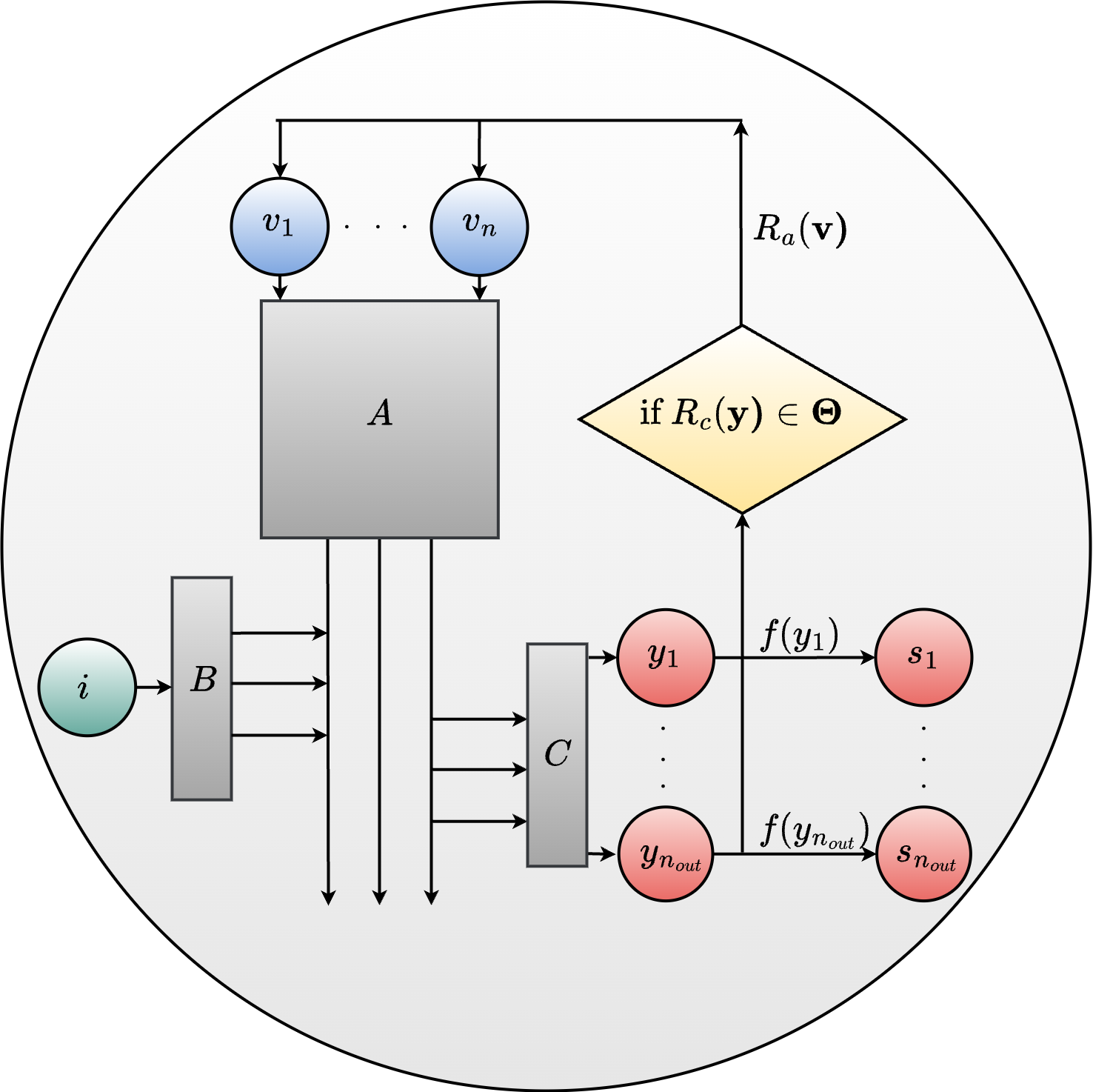}
    \caption{Illustration of  dynamics of a neuron with multiple-outputs  and a general reset mechanism.}
    \label{fig:SIMO_single_neuron_dynamics}
\end{figure}

% Reset
Here, $\Thetabf$ defines a region in the range space of $\RstFncC(.)$.
At each time step,  if the reset condition, $\RstFncC(\ybf[t]) \in \Thetabf$ based on the output $\ybf$, is satisfied then the reset action, $\vbf[t+1] =\RstFncA(\vbf[t+1])$, is applied to the state. See Figure \ref{fig:SIMO_single_neuron_dynamics} for an illustration.
    
\subsubsection{Spiking function}
\label{sec:method:spk_fnc}
In this paper, as the spiking function, $\ActFnc_\theta(\cdot)$, we use the most common spiking function -- the Heaviside step function with threshold $1$. However, in most of the existing works in SNN literature,  the spiking function is usually applied to the state variables and furthermore, it is typically applied  only to one of the state variables which corresponds to the membrane potential. Here,  we make a distinction between the state variable $\vbf$ and the output of the neuron $\ybf$,  and apply the spiking function to the output  $\ybf$ of the neuron rather than $\vbf$. In particular, we  apply $\ActFnc_\theta(\cdot)$  element-wise to $\ybf$ such that for $i=1,...,\nout$ we have
\begin{align}
\label{eqn:spikingFunction:elementwise}
\ActFnc_\theta(\ybfi[t])
&= \begin{cases} 
   1  \text{ if } \Re(\ybfi[t])+\Im(\ybfi[t]) > 1 \\
   0  \text{ otherwise.}
   \end{cases}
\end{align}
This formulation allows the neuron to produce $\nout$ outputs regardless of the state dimension.   

At first glance, \eqref{eqn:spikingFunction:elementwise} seems to impose a spiking threshold of $1$ for all output channels of the neuron. 
We recall that the model includes trainable $\Cbf$ and $\CbfBias$,  hence different outputs can in fact have different spiking thresholds. This feature allows diversity in the output and may be used to compensate for the negative effect of the  1-bit amplitude resolution in each output channel of the neuron. In particular,  a multiple output neuron with different learnable spiking thresholds for each channel can be semantically interpreted as a single-output neuron with   multiple thresholds and spike levels, e.g. a neuron that can output  graded spike events as in Loihi 2 \cite{davies_advancing_2021, meyer2024diagonalstructuredstatespace}. Hence, the multiple-outputs  of \eqref{eqn:general_spiking_ssm_neuron:SIMO} may be interpreted effectively as a projection of the amplitude resolution of a graded spike into space, i.e., onto different channels in the output, with the added benefit of using different, possible learnable parts of the output space for different channels through learnable $\Cbf$. The positive effect of increasing $\nout$, and hence possible diversity, is illustrated in our numerical results in Section~\ref{sec:results:mswc-H_vs_N_vs_Nout}.

\subsubsection{Reset Condition}
In biologically inspired computational neuron models, the reset condition $\RstFncC(\cdot)$ is typically linked to the spiking function, meaning that a reset occurs whenever the neuron emits a spike. However, in this work, we introduce a key distinction: the spiking and reset mechanisms operate independently, allowing for separate dynamics.

We further differentiate between hard and soft reset conditions. Instead of solely indicating whether a reset should occur, a soft reset can provide quantitative information about the reset magnitude. However, in this work \eqref{eqn:general_spiking_ssm_neuron:SIMO:reset}, we employ a hard reset condition, where the only information transmitted is a binary decision on whether to reset or not.

In this paper, we consider a hard reset condition but decoupled from the spiking mechanism. Specifically, the reset function $\RstFncC(\cdot)$ is defined using the Heaviside step function, combined with the Euclidean norm and a learnable bias term $\RstBias \in \R^{1 \times 1}$ as follows: 
\begin{align}
    \label{eqn:reset_condition:explicit}
    \RstFncC(\ybf[t]) \in \text{$\Thetabf$}  
    \equiv
    \frac{1}{\nout}\|\ybf[t]\|_2+\RstBias \geq 1
\end{align}

\begin{figure}
    \centering
    % subfigure
    \begin{subfigure}{0.49\linewidth}
        \centering
        \includegraphics[width=\linewidth]{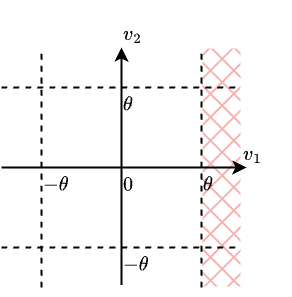}
        \caption{adLIF neuron reset condition with $v_1>\theta$.}
        \label{fig:spk_region_adLIF}
    \end{subfigure}
    % subfigure
    \begin{subfigure}{0.49\linewidth}
        \centering
        \includegraphics[width=\linewidth]{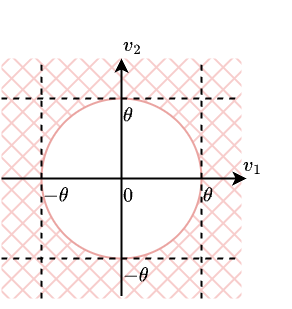}
        \caption{Proposed neuron reset condition for the  case of $\| \vbf \|_2 \geq \theta$.}
        \label{fig:spk_region_SSM}
    \end{subfigure}
    \caption{Visualization of the reset condition region, represented by red cross hatch, for $\Nstate=2$ dimensional neurons.}
    \label{fig:spk_region}
\end{figure}

Note, in an adLIF neurons with state variables $\vbf=\begin{bmatrix} v_1, v_2 \end{bmatrix}^T \in \R^{2 \times 1}$, the reset condition is triggered when the membrane potential exceeds a threshold i.e.,  $v_1 \geq \theta$, see the reset condition region in Figure~\ref{fig:spk_region_adLIF}. On the other hand, the proposed reset condition is possibly based on all state variables through $\ybf$. Consider the following example:  Let $\vbf \in  \R^{2 \times 1}$, $\Cbf=2 \Ibf \in \R^{2 \times 2}$, $\CbfBias=0$, $\RstBias =1-\theta$. Then the reset condition becomes $\RstFncC(\ybf) =\| \vbf \|_2 \geq \theta$, see Figure~\ref{fig:spk_region_SSM} for an illustration.

\begin{remark}
In most commonly used spiking neuron models in SNNs, the reset condition coincides with spike generation, i.e., the reset condition is met when the neuron emits a spike. We generalize this by decoupling the two processes, using distinct functions, $\ActFnc_\theta(\cdot)$ for spike generation and $\RstFncC(\cdot)$ for the reset condition.  Furthermore, in most neuron models used in SNNs,   the spiking depends only on one of the state variables; and hence  also the reset condition. In contrast, our approach defines the reset condition based on a multi-output function that incorporates all state variables.
\end{remark}

\subsubsection{Reset Action}
\label{methods:reset_action}
In this paper, we model the reset action, $\RstFncA(\cdot)$, using a simple element-wise scaling of the state variables $\vbfi$, $k=1,...,\Nstate$ with a learnable scaling parameter $\RstValue \in \C^{1 \times 1}$. Hence, \ref{eqn:general_spiking_ssm_neuron:SIMO:reset} becomes 
\begin{align}
\label{eqn:general_spiking_ssm_neuron:SIMO:reset:explicit}
\! \vbfi[t+1] &= \begin{cases} 
          \RstValue \vbfi[t+1]  \text{ if Reset Condition is True}\\
          \vbfi[t+1]  \text{ otherwise}
          \end{cases}
\end{align} 

We now compare the reset mechanism in \eqref{eqn:general_spiking_ssm_neuron:SIMO:reset:explicit} with typical reset mechanisms in neuromorphic computing. 
In various biologically inspired neuron models, e.g. LIF, the reset condition typically affects only the membrane potential, which is one of the state variables. This process is commonly referred to as the {\it{postsynaptic potential (PSP)}}. In contrast to these models, here we propose a generalized change in all state variables when reset occurs. In some models, e.g. LIF,  the reset mechanism involves subtracting the threshold value $\theta$ from the membrane potential to equate the energy lost from the output spike  \cite[Eqn.7]{bittar2022surrogate} \cite[Eqn.3]{neftci2019surrogate}. In contrast, here we consider a generalized reset action that is not directly connected to the threshold for the spiking function. Some models incorporate a {\it{refractory period}} \cite{Zenke_2018, ZHAO202280, 9816777}, imposing constraints on the behavior of membrane potential after a spike for a certain period. Instead, in \eqref{eqn:general_spiking_ssm_neuron:SIMO:reset:explicit} an immediate reset action is adopted and the neuron continues to follow the original neuron model.

\begin{remark} 
In the existing popular SNN models, such as LIF and adLIF, the reset action is coupled to the spiking function such that whenever there is a spike out, it gets subtracted from one of the state variables associated with the membrane potential. In our work, we go a step further by introducing trainable reset parameter associated with how much all of the state variables would be scaled.
\end{remark}

\subsubsection{Neuron Model to Network}\label{sec:method:proposed:network}

\begin{figure}
    \centering
    \includegraphics[width=0.49\textwidth]{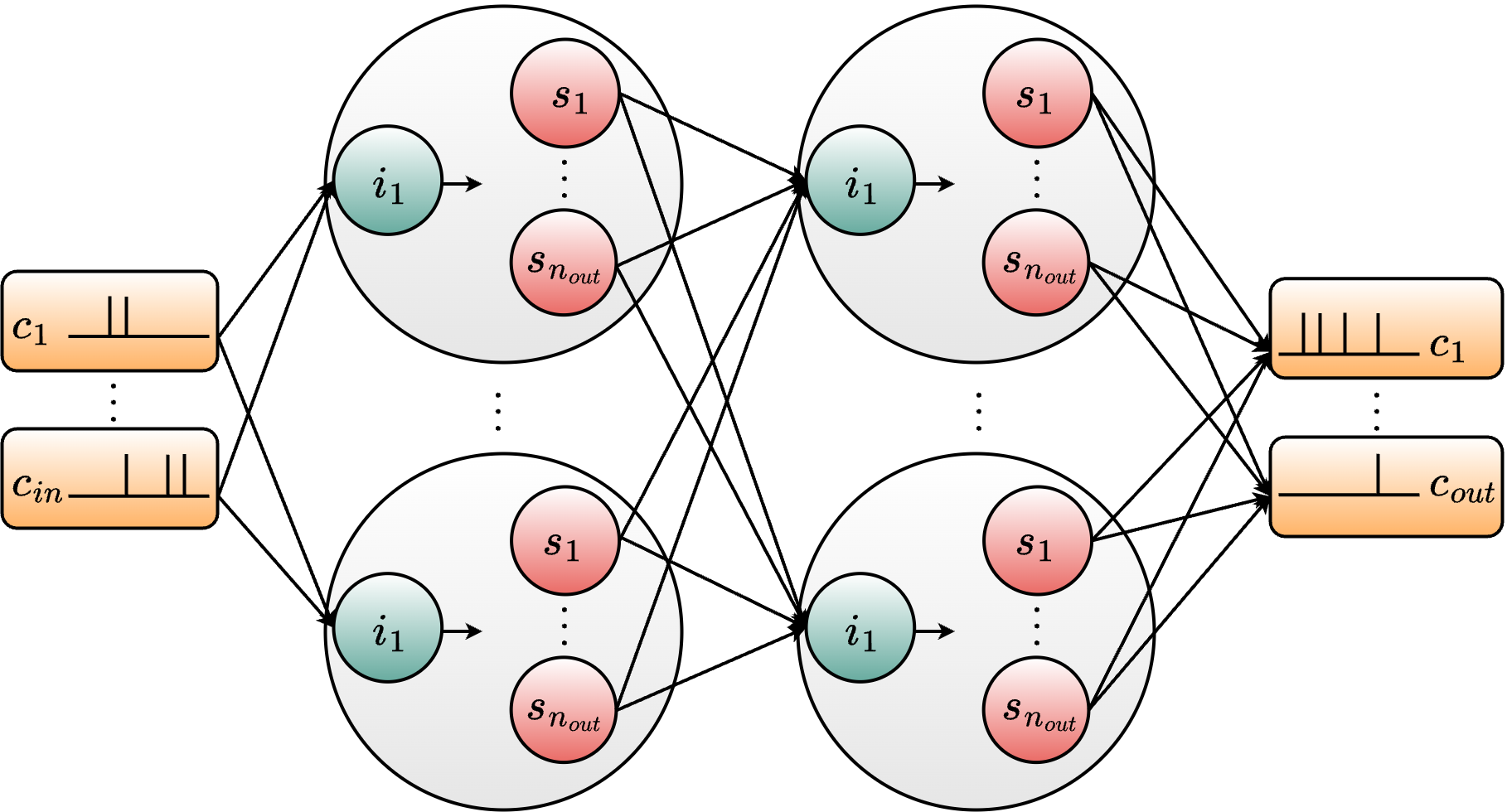}
    \caption{An illustration of a network with two hidden layers and multiple-output spiking neurons.}
    \label{fig:SIMO_network}
\end{figure}

Following the spiking neural networks framework, connections between stateful hidden layers with multiple-output neurons is constructed by connecting output channels across neurons using a dense weight matrix $\Wbf \in \R^{\Hneurons \nout \times \Hneurons}$. This weight matrix parametrizes the connections between  spiking outputs from the previous pre-synaptic layer, i.e., $\Sl{t}$'s,  to the inputs of the next post-synaptic layer, i.e., $\Inbf[t]$'s. See Figure \ref{fig:SIMO_network} for an illustration. We note that Figure \ref{fig:SIMO_network} provides a high-level illustration of a network,  including the input  and output channels of the network. In our experiments, possibly different number of hidden layers are used depending on the dataset. The network structures are discussed in Section \ref{sec:network_architecture}.

\subsection{Computational Cost}
\label{sec:computational-cost-general}

We now discuss the additional computational cost due to the proposed reset mechanism, focusing on the learnable parameters in Section~\ref{sec:number-of-params} and number of multiply–accumulate (MAC) operations  in Section~\ref{sec:number-of-operations}.   

\subsubsection{Number of learnable parameters}
\label{sec:number-of-params}

We now discuss the number of learnable parameters in our proposed neuron model in Section \ref{sec:proposed_model}.
Although $\Abf$ in \eqref{eqn:general_spiking_ssm_neuron:SIMO:state} can be parametrized as a possibly dense  trainable matrix, motivated by  the success of models with diagonal state transition matrices \cite{gu2022parameterization, meyer2024diagonalstructuredstatespace, Gupta_DSS} and our own numerical results in Section~\ref{sec:results},  here we focus on the case $\Abf$ is diagonal. 
Therefore, we have $\Nstate$ complex-valued parameters for diagonal $\Abf$.
For  $\Bbf$ in \eqref{eqn:general_spiking_ssm_neuron:SIMO:state} we have  $\Nstate$ real-valued parameters. For $\Cbf$ and $\CbfBias$ in \eqref{eqn:general_spiking_ssm_neuron:SIMO:obsy} we have $\Nstate \times \nout$ complex-valued parameters, and  $\nout$ complex-valued parameters, respectively. 
There are no trainable parameters in \ref{eqn:general_spiking_ssm_neuron:SIMO:obsspike}. 
For \ref{eqn:general_spiking_ssm_neuron:SIMO:reset}, we have $1$ real-valued  parameter ($\RstBias$) from reset condition \eqref{eqn:reset_condition:explicit}  and 
$1$ complex-valued parameter ($\RstValue$) from reset action \eqref{eqn:general_spiking_ssm_neuron:SIMO:reset:explicit}. 
Therefore, accounting for the fact that one complex-valued parameter corresponds to two real-valued parameters,  our proposed neuron model in Section \ref{sec:proposed_model} has an equivalent of $p =(2 \nout + 3)(\Nstate + 1)$ trainable real-valued parameters.
Hence, there are $p \times\Hneurons$ trainable parameters per hidden layer with $\Hneurons$ neurons. 

% Comparison to NoReset
The above analysis demonstrates that the reset mechanism adds a small overhead in terms of trainable parameters, introducing only three additional trainable parameters per neuron.  In the context of typical values of $\Nstate$ and $\nout$ in SSM-based deep models and also in our models, see Section~\ref{sec:appendix:numerical:settings_preliminaries:training}, this overhead is not significant.

\subsubsection{Number of operations}
\label{sec:number-of-operations}

We now focus on the additional operations introduced by the reset mechanism per time step.

Evaluating the reset condition \eqref{eqn:reset_condition:explicit} requires a total of $m_{rc}=4\nout+1$ MACs. Specifically,  focusing on multiplications/additions, we have $2 \nout$ real multiplications (i.e. multiplication of real-valued parameters), $2 \nout-1$ real  additions  for calculation of   $\|\ybf[t]\|_2$, 1 multiplication when scaling with $\frac{1}{\nout}$, 1 addition with $\RstBias$. When triggered, the reset action of \eqref{eqn:general_spiking_ssm_neuron:SIMO:reset:explicit} scales the state variable, requiring $\Nstate$ complex multiplications, each corresponding to 4 real multiplications and 2 real additions, hence $m_{ra}= 6\Nstate $ MACs. Hence, assuming a reset occurs, the reset mechanism introduces $m_r =m_{rc}+m_{ra} =4\nout+6 \Nstate+1$ MACs per neuron, and $m_r \times \Hneurons$ MACs per hidden layer with $\Hneurons$ neurons.

The computational cost of the reset mechanism is typically negligible compared to the computational cost related to the synaptic weights for a fully connected network architecture. In particular, multiplication with the weight matrix for the synaptic connections from  one hidden layer to another hidden layer requires  MACs of the order of $\Hneurons \times \nout \times \Hneurons$. As illustrated in Section~\ref{sec:appendix:numerical:settings_preliminaries:training}, $\Hneurons$ is typically significantly larger than both $\nout$ and $\Nstate$ in our models.  Hence, focusing on the scaling with $\Hneurons$, the hidden layer connections require  $\bigO(\Hneurons^2)$ operations per time step due to weight matrix operations, whereas the reset mechanism adds only $\bigO(\Hneurons)$ operations. Therefore, the contribution of the reset mechanism to the total computational load is relatively insignificant.

\section{Experimental Setup}
\label{sec:settings_preliminaries}

In this section, we present the set-up for numerical results. This section provides an overview of the main experimental design choices.  Whenever applicable,  further details are provided in the Appendix in Section~\ref{sec:appendix:numerical:settings}.

\subsubsection{Neuron Models}\label{sec:num:neuronModels}
We investigate four neuron model structures based on two key factors: 
the presence of a reset mechanism and stability of neuron state dynamics. 
For the reset mechanism, we have the following two cases: 
\begin{itemize}
\item The neuron resets, 
i.e., it follows \eqref{eqn:general_spiking_ssm_neuron:SIMO:state}-\eqref{eqn:general_spiking_ssm_neuron:SIMO:reset}
\item The neuron does not reset, i.e, it follows \eqref{eqn:general_spiking_ssm_neuron:SIMO:state} -\eqref{eqn:general_spiking_ssm_neuron:SIMO:obsspike}. 
\end{itemize}
For the linear part of the neuron state dynamics, we have the following two cases:  
\begin{itemize}
\item Stable: $\Abf$ is initialized so that the initial neuron dynamics is stable. During training, the eigenvalues are clipped to ensure stability is maintained. 
\item Unstable: $\Abf$ is initialized so that the initial neuron dynamics is possibly unstable. During training, eigenvalues are not clipped, hence unstable dynamics is allowed.  
\end{itemize}

According to the above, we consider four different neuron models in our experiments: \stablereset, \stablenoreset, \unstablereset, and  \unstablenoreset. Note that we refer to the linear part of dynamics of the neuron model, i.e. \eqref{eqn:general_spiking_ssm_neuron:SIMO:state},  while using the label {\it{stable}} or {\it{unstable}}. Note that the stability properties of the overall neuron dynamics may change due to the reset mechanism.  

In all experiments, diagonal state transition matrices with learnable eigenvalues are used. The state transition matrices is then $\Abf=\Lambdabf$ where $\Lambdabf$ is a diagonal matrix of learnable eigenvalues.
Further details for the neuron model,  initializations and clipping operations  are provided in
Section~\ref{sec:appendix:numerical:settings_preliminaries:neurons}.

\subsubsection{Network Architecture}
\label{sec:network_architecture}

% Basics
The number of input channels of a network is determined by the number of data sequences in the dataset. The number of output channels is determined by the number of classes in the classification task.  
The input and output layers, i.e.  the layers consisting of the input and output channels, are state-less. 
In between the input and output layers, stateful hidden layers whose number depends on the dataset are included, see Figure~\ref{fig:SIMO_network} for a visualization with two hidden layers. The network architecture parameters for different datasets are provided in Section~\ref{sec:appendix:numerical:settings}. 
We have trainable synaptic weights between the input layer and the first hidden layer, between the hidden layers, and between the last hidden layer and the output layer.
Our networks include batch normalization between layers \cite{ioffe2015batchnormalizationacceleratingdeep}.

% Decisions
In classification tasks, which are the focus of this paper, decisions are made based on the accumulated output spikes. In particular,  each output channel corresponds to a specific class, and the  class estimated by the network is given by the class with the highest spike count, i.e.,  rate-based decoding.

\subsubsection{Activation functions}
\label{sec:act_fnc}

\begin{figure}
    \centering
    % subfigure
    \begin{subfigure}{0.95\linewidth}
        \centering
        \includegraphics[width=\linewidth]{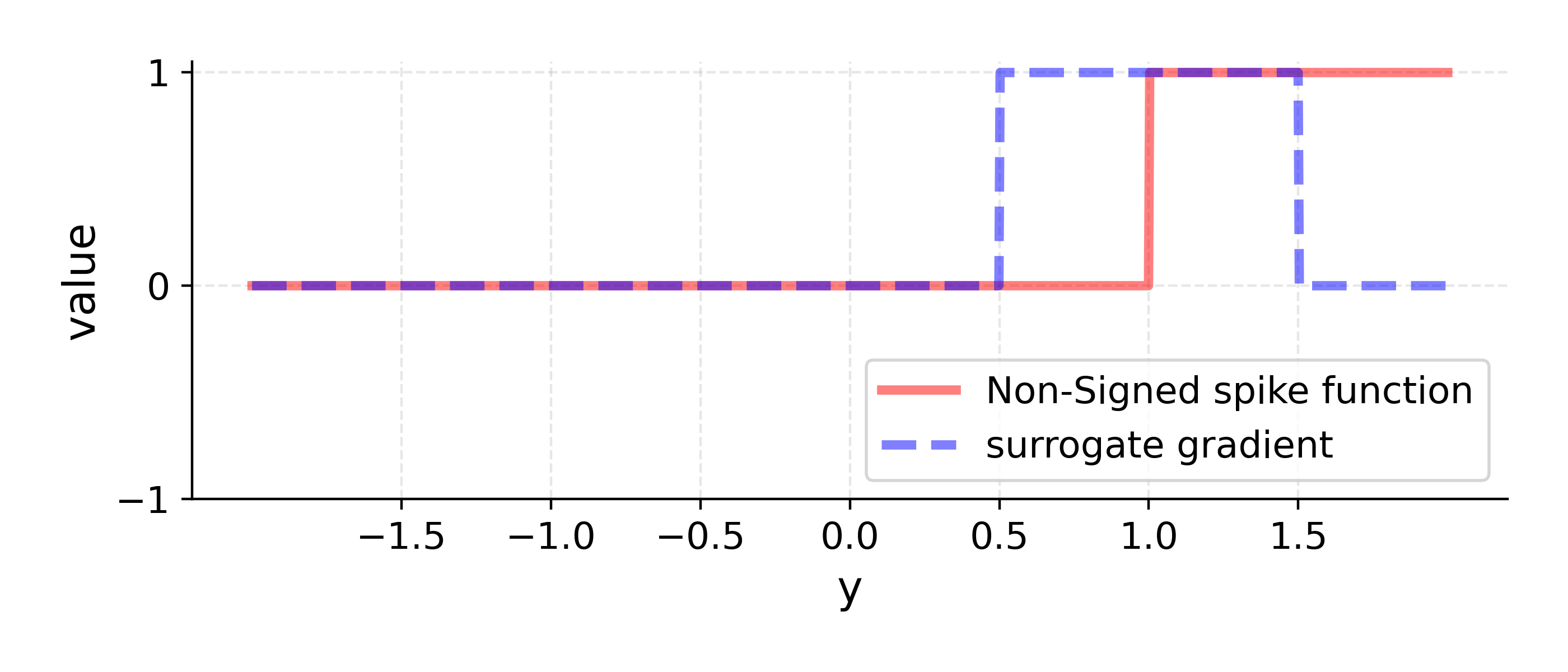}
        \caption{{\it{Non-Signed spike}} activation function}
        \label{fig:nonsigned_spike}
    \end{subfigure}
    % subfigure
    \begin{subfigure}{0.95\linewidth}
        \centering
        \includegraphics[width=\linewidth]{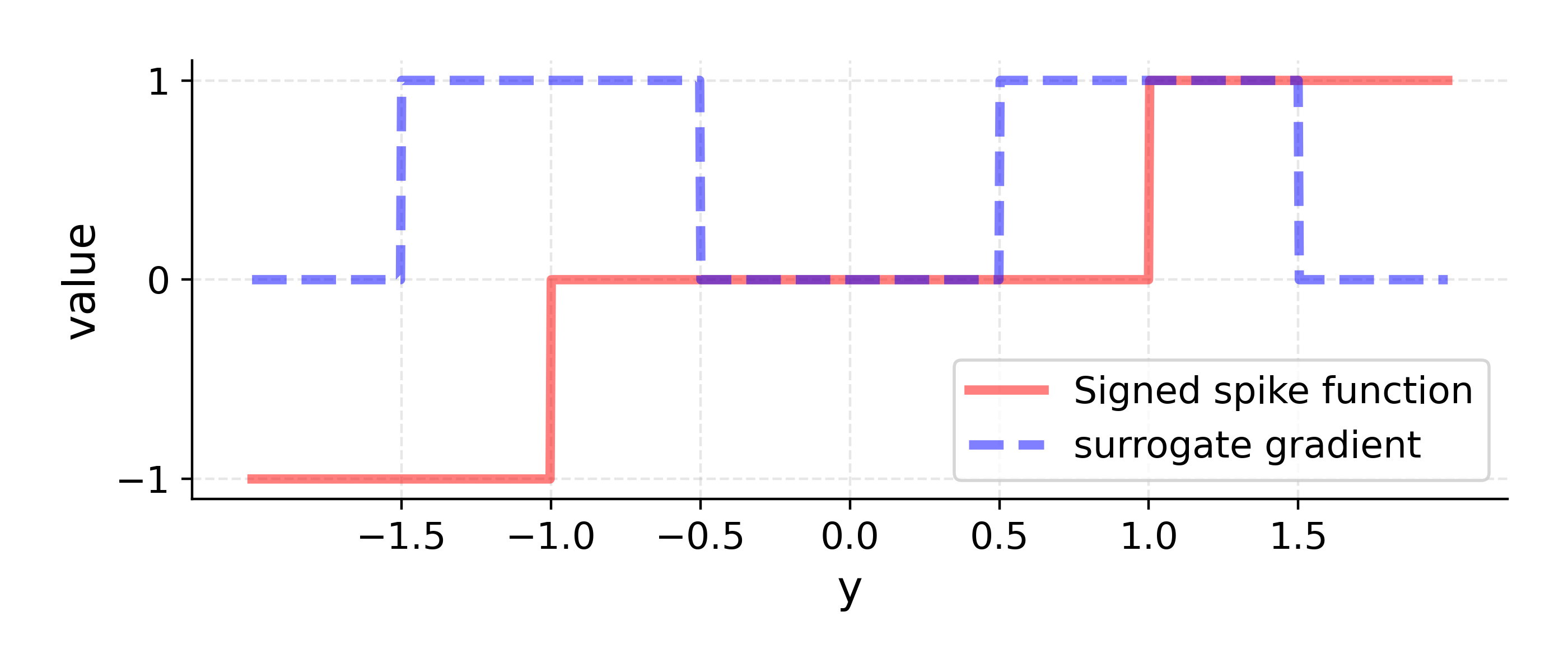}
        \caption{{\it{Signed spike}} activation function}
        \label{fig:signed_spike}
    \end{subfigure}
    \caption{Visualization of activation function and their surrogate gradient for spiking threshold=1}
    \label{fig:spike_and_SG}
\end{figure}

% Intro
The main interest of this paper is binary valued $\{0,1\}$ information transfer between neurons, i.e. output of the function $\ActFnc_\theta(\cdot)$ in \eqref{eqn:general_spiking_ssm_neuron:SIMO:obsspike} is binary-valued. More specifically,  we use the {\it{non-signed spikes}} (unipolar) illustrated in Figure~\ref{fig:nonsigned_spike}. We also explore a 
{\it{signed spikes}} (bipolar)  spiking function with ternary output values $\{-1,0,1\}$. This function  visualized in Figure!\ref{fig:signed_spike}. Compared to the {\it{non-signed spikes}}, {\it{signed spikes}}  increase  the level of quantization from $1$-bit to $2$-bit by including a sign. For both activation functions,  surrogate gradient \cite{neftci2019surrogate} of car-box function at each spiking point is used in the backward pass as approximation to the otherwise non-differentiable activation function, see Figure~\ref{fig:nonsigned_spike} and Figure!\ref{fig:signed_spike}. For benchmarking purposes, i.e. in order to quantify the possible performance loss due to low-bit representation with spikes, performance of   the continuous-valued {\it{GELU}} activation function is also provided  in some experiments.

% Signed-spike surrogate gradient

\subsubsection{Training}
\label{sec:network_training}
The models are trained using cross-entropy loss and Batch-propagation through time (BPTT)  where surrogate gradient  \cite{neftci2019surrogate, bittar2022surrogate} is used for the two heaviside step functions in our approach -- the spiking condition of neurons, and the reset condition. 
Further details are provided in Section~\ref{sec:appendix:numerical:settings_preliminaries:training}.

\subsubsection{Datasets}

\paragraph{Spiking audio processing}

% MSWC
The Multilingual Spoken Word Corpus (MSWC) dataset \cite{mswcdataset} is a large and growing dataset consisting of spoken words in over $50$ languages. In this paper we use a subset of the dataset and its encoding as proposed in the NeuroBench initiative \cite{yik2024neurobench}. 
The subset consists of $100$ classes representing different spoken words featuring $20$ words per $5$ different languages, see Section~\ref{sec:appendix:datasets:mswc} for further information.
The aim is to classify these $100$ words using audio data that is represented using spikes. 
The current baseline performance for classification accuracy on this subset is $97.1\%$ using continuous-valued ANN \cite{yik2024neurobench}.

\paragraph{Event-based vision processing}

The DVS-Gesture dataset \cite{Amir_2017_CVPR} is one of the most popular benchmarks datasets for event-based vision processing. It consists of $29$ subjects performing $11$ hand gestures under $3$ different lighting conditions. The dataset is recorded using dynamic vision sensor of $128 \times 128$ resolution with two channels encoding the two event polarities.
See Section~\ref{sec:appendix:datasets:dvs_gesture} for further information.
The aim is to classify the $11$ hand gestures. The current baseline performance for classification accuracy on this dataset is $99.3 \%$ using a transformer based architecture \cite{yao2023spike} and $97.7\%$  with a   SSM-based architecture \cite{schöne2024scalableeventbyeventprocessingneuromorphic}.

\paragraph{Sequential pattern recognition}

sMNIST dataset \cite{le2015simple} is a popular sequential version of the  MNIST dataset \cite{mnist_dataset},
providing a popular benchmark for evaluating sequential processing.
Each $28\times28$ image of MNIST corresponds to a  handwritten digit, i.e. $0$ to $9$.
In sMNIST, each image is flatten into a time sequence of length $28\times28=784$. Hence instead of processing all pixels at once as an image, the sample is processed as a time sequence as one pixel at each time step.  
See Section~\ref{sec:appendix:datasets:sMNIST} for further information.
Similar to MNIST, the aim is to classify the $10$ digits.
The current baseline performance for classification accuracy on this dataset is
$99.65 \%$ using continuous-valued S5 \cite{smith2023simplified}.

\section{Results}
\label{sec:results}

% Content
In the following sections, we investigate various aspects that influence model behavior and performance. 
Section~\ref{sec:results:overview} provides an overview of the results. 
Section \ref{sec:results:impact_of_reset_on_stable_or_unstable} examines the impact of reset mechanisms under stable and unstable dynamics. Section \ref{sec:results:activation_functions_comparisons} explores how low-bit resolution communication affects different neuron models. Training dynamics—specifically stability and convergence—are studied in Section \ref{sec:results:Training_over_epochs}. Section \ref{sec:results:Training_over_time} focuses on time-to-decision behavior, i.e., accuracy progression over time. Section \ref{sec:results:spike_rate} provides insights into the spiking rates of the models. Section \ref{sec:results:Drop_Output_channels} evaluates the importance and redundancy of multiple output channels. Finally, Section \ref{sec:results:mswc-H_vs_N_vs_Nout}, analyzes accuracy trade-offs across different combinations of $\Hneurons$, $\Nstate$, and $\nout$ under architectural constraints for a specific dataset.

% General
All tables report classification accuracy on the test dataset as a percentage, where $100\%$ indicates perfect classification on the test set. These results correspond the test accuracy at the final epoch, and represent the mean accuracy over $5$ runs with different random initializations of the network. Standard deviations across these runs are also included.

\begin{table*}[]    
    \centering
    \caption{{Classification accuracy (\%) for all three datasets for different models with varying the neuron dynamics and amplitude resolution of the activation function. Related literature is grouped into neuroscience-inspired and deep-state-space-inspired, separated with dashed-line. The highest accuracy in our models is in bold; the best from the related work is both bolded and boxed. All accuracy values are rounded to $1$ decimal points. Number of parameters for the models in the literature is reported when the original work provides this information. Table \ref{tab:hyperparams} provides the model hyperparameters.}} 
    \label{tab:All_Datasets:model_vs_ActFnc}
    \newcolumntype{l}{>{\centering\arraybackslash}p{2.7cm}}
    \newcolumntype{n}{>{\centering\arraybackslash}p{1.6cm}}
    \newcolumntype{k}{>{\centering\arraybackslash}p{2cm}}
    \newcolumntype{m}{>{\raggedright\arraybackslash}p{2.2cm}}
    
    \begin{subtable}[h]{1\linewidth}
    \begin{threeparttable}
    \centering
    \caption{MSWC dataset.}
    \vspace{3pt}
    \label{tab:mswc:model_vs_ActFnc}
    %----------------------------------------------------
    \begin{tabular}{lnkmmmn}
    \hline
        Model &  Dynamics & Reset
               & {\it{Non-signed spikes}}
               & {\it{Signed spikes}}
               & {\it{GELU}} 
               & Num. params. \\
    \hline
    Ours & Stable & Yes -- Proposed  &\textbf{95.0 $\pm$ 0.1} \% & \textbf{95.1 $\pm$ 0.1} \% & 95.6 $\pm$ 0.1 \% &0.4M\\
    Ours & Stable & No &88.9 $\pm$ 0.4 \% & 92.1 $\pm$ 0.2 \% & \textbf{95.7 $\pm$ 0.0} \% &0.4M\\
    Ours & Unstable & Yes -- Proposed  &91.5 $\pm$ 0.4 \% & 93.9 $\pm$ 0.5 \% & 94.2 $\pm$ 0.3 \% &0.4M\\
    Ours & Unstable & No &40.2 $\pm$ 1.1 \% & 50.7 $\pm$ 1.1 \% & 90.6 $\pm$ 0.4 \% &0.4M\\
    \hdashline
    RadLIF \cite{yik2024neurobench} & Stable & Yes -- LIF & 93.5 \% & - &  - & 3.4M{\tnote{(a)}} \\
    RadLIF \cite{karilanova2024zeroshottemporalresolutiondomain} & Stable & Yes -- LIF & \boxit{2.2cm}\textbf{95.7 \%} & - &  - & 3.3M \\
    \hdashline
    M5 ANN \cite{yik2024neurobench} & N/A & No & - & - &  \boxit{2.2cm}\textbf{97.1} \% & 0.5M{\tnote{(b)}}, \,\,  1.5M{\tnote{(a)}}  \\
    \hline
    \end{tabular}
    \begin{tablenotes}
    \small
        \item[(a)] \cite[Table~2]{yik2024neurobench},  calculated as num.~params $\approx$ (Footprint in bytes)/4. 
         \item[(b)]  \cite[Table~1]{Dai_2017}.
    \end{tablenotes}
    %----------------------------------------------------
    \end{threeparttable}
    \end{subtable}

    \begin{subtable}[h]{1\linewidth} 
    \centering
    \caption{DVS-Gesture dataset.}
    \vspace{3pt}
    \label{tab:dvs:model_vs_ActFnc}
    %----------------------------------------------------
    \begin{tabular}{lnkmmmn}
    \hline
        Model &  Dynamics & Reset
               & {\it{Non-signed spikes}}
               & {\it{Signed spikes}}
               & {\it{GELU}} 
               & Num. params. \\
    \hline
    Ours & Stable & Yes -- Proposed &\textbf{93.0 $\pm$ 1.8} & 94.0 $\pm$ 1.3 & \textbf{97.1 $\pm$ 0.3}& 5.2M\\
    Ours & Stable & No &90.1 $\pm$ 3.0 & \textbf{95.4 $\pm$ 0.8} & 92.4 $\pm$ 3.8& 5.2M\\
    Ours & Unstable & Yes -- Proposed &69.0 $\pm$ 7.0 & 84.0 $\pm$ 2.2 & 95.7 $\pm$ 0.4& 5.2M\\
    Ours & Unstable & No &75.4 $\pm$ 14.0 & 83.5 $\pm$ 2.7 & 95.4 $\pm$ 1.3& 5.2M\\
    \hdashline
    DECOLLE \cite{10.3389/fnins.2020.00424} & Stable & Yes-LIF & 95.5 \% &-& - &  - \\
    tdBN \cite{Zheng_Wu_Deng_Hu_Li_2021} & Stable & Yes-LIF &96.9 \%&-& - &  - \\
    PLIF \cite{Fang_2021_ICCV} & Stable & Yes-LIF &\boxit{2.2cm}\textbf{97.6} \%&-& - &  - \\
    \hdashline
    EventSSM \cite{schöne2024scalableeventbyeventprocessingneuromorphic} & Stable & No &-&-&\boxit{2.2cm}\textbf{97.7} \% & 5M \\
    \hline
    \end{tabular}
    %----------------------------------------------------
    \end{subtable}

    \begin{subtable}[h]{1\linewidth} 
    \begin{threeparttable}
    \centering
    \caption{sMNIST dataset.}
    \vspace{3pt}
    \label{tab:smnist:model_vs_ActFnc}
    %----------------------------------------------------
    \begin{tabular}{lnkmmmn}
    \hline
        Model &  Dynamics & Reset
               & {\it{Non-signed spikes}}
               & {\it{Signed spikes}}
               & {\it{GELU}} 
               & Num. params. \\
    \hline
    Ours & Stable & Yes -- Proposed  &95.8 $\pm$ 0.5 \% & 96.2 $\pm$ 0.5 \% & 95.9 $\pm$ 0.6 \% & 113k \\
    Ours & Stable & No &\textbf{98.8 $\pm$ 0.1} \% & \textbf{99.1 $\pm$ 0.1} \% & \textbf{99.3 $\pm$ 0.1} \% & 113k \\
    Ours & Unstable & Yes -- Proposed &96.3 $\pm$ 0.4 \% & 97.3 $\pm$ 0.4 \% & 97.2 $\pm$ 0.3 \% & 113k \\
    Ours & Unstable & No &82.9 $\pm$ 0.4 \% & 84.8 $\pm$ 0.2 \% & 95.2 $\pm$ 0.2 \% & 113k \\
    \hdashline
    LIF \cite{moro2024role} & Stable & Yes -- LIF  & 89.5 \% \tnote{(d)} &-& - & -\\
    RLIF \cite{moro2024role} & Stable & Yes -- LIF  &  96.9  \% \tnote{(d)} &-& - & -\\
    RadLIF \cite{shaban2021adaptive} & Stable & Yes -- LIF  &96.1 \% \tnote{(c)}&-&  & -\\
    \hdashline
    S4D \cite{meyer2024diagonalstructuredstatespace} & Stable & No &-&-& 99.5 \% \tnote{(a)} & 67k\\
    S5 \cite{smith2023simplified} & Stable & No &-&-&\boxit{2.2cm}\textbf{99.7} \% & -\\
    Q-S5  \cite{abreu2024qs5quantizedstatespace} & Stable & No &-&-&99.6 \% \tnote{(b)} & - \\
    SpikingSSM \cite{shen2024spikingssmslearninglongsequences} & Stable & Yes -- LIF output & \boxit{2.2cm}\textbf{99.6} \% &-&-& - \\
    BinarySSM \cite{stan2024learning} & Stable & Yes -- LIF output & 99.4 \% &-&-& 118k \\
    \hline
    \end{tabular}
    \begin{tablenotes}
    \small
        \item[(a)] RELU instead of GELU used. 
        \item[(b)] 8-bit quantized GELU (qGELU) used. 
        \item[(c)] Reported best accuracy run.
        \item[(d)] Read from \cite[Fig. S5]{moro2024role}.
    \end{tablenotes}
    %----------------------------------------------------
    \end{threeparttable}
    \end{subtable}
\end{table*}

\subsection{Overview}\label{sec:results:overview}

% Comparison to SoTA
In Table~\ref{tab:All_Datasets:model_vs_ActFnc} we compare the four neuron models of our framework under different activation functions.  The table also presents relevant results from the literature, covering both neuroscience-inspired SNNs and models based on the recent deep state-space formulations. We observe that under both spike-based communication  and continuous-valued communication, our models achieve accuracy comparable to state-of-the-art methods while maintaining reasonable and even smaller model sizes across all datasets. 
For instance, our models with the proposed reset mechanism achieve accuracy values close to or matching the best reported results with non-signed spikes as follows:  
$95.0 \%$ ({\stablereset}) vs.  $95.7\%$   (RadLIF of   \cite{karilanova2024zeroshottemporalresolutiondomain}) on MSWC, 
$93.0\%$ (\stablereset) vs. $97.6\%$   (PLIF \cite{Fang_2021_ICCV}) on DVS-Gesture, 
and $96.3$ \% ({\unstablereset}) vs. $99.6\%$  (SpikingSSM \cite{shen2024spikingssmslearninglongsequences}) on sMNIST.

% Generalize across modalities
These results indicate that our proposed model generalizes well across modalities, including audio,  vision and sequential pattern recognition tasks. 
These results also illustrate that  our proposed model can handle relatively long-sequences as input. 
Notably, DVS-Gesture (with an average of $\approx 340$ time steps, see Section~\ref{sec:appendix:datasets:dvs_gesture}) and sMNIST (with $784$ time steps) has samples with relatively long-sequences in comparison to many other standard datasets in the neuromorphic computing literature, such as the Spiking Heidelberg Digits (SHD) dataset and Spiking Speech Command (SSC), which are often used with $100$ time steps \cite{bittar2022surrogate, moro2024role, fabre2025structuredstatespacemodel}.

All subsequent results use our models from Table~\ref{tab:All_Datasets:model_vs_ActFnc} unless otherwise stated explicitly.

\subsection{Impact of reset under stable and unstable dynamics}
\label{sec:results:impact_of_reset_on_stable_or_unstable}

We now focus on Table~\ref{tab:All_Datasets:model_vs_ActFnc}  and compare the four neuron models ({\stable} or {\unstable}, {\reset} or {\noreset}) under our proposed framework.

% If A is stable, reset vs no reset?
For {\stable} dynamics, how the reset affects the performance depends on the dataset and the activation function.
For example, under {\it{non-signed spikes}},  for the MSWC dataset and DVS-Gesture dataset it increases the accuracy from $88.9\%$ to $95.0\%$ and $90.1\%$ to $93.0\%$, respectively. 
On the other hand, for the sMNIST dataset, it drops the performance from  $98.8\%$ to $95.8 \%$.

% If A is initialized unstable, it is crucial to use reset?
For {\it{Unstable}} dynamics, using reset  increases the performance for MSWC and sMNIST.  The performance improvement is large for the low bit communication cases of {\it{non-signed and signed spikes}} whereas it is relatively smaller under the continuous-valued communication case of GELU. For example,  an increase from $40.2\%$ to $91.5\%$ for MSWC, and $82.9\%$ to $96.3\%$ for sMNIST is observed under {\it{non-signed spikes}}. 
For DVS-Gesture, the performance changes are within standard deviation limits where the performance has  large fluctuations under  {\it{non-signed spikes}}, suggesting further regularization and reset optimization may be useful.

We note the results in the table for unstable dynamics without reset are obtained with clipping of the state and clipping of gradients.  Without clipping, the computation cannot be performed in this scenario due to diverging states and exploding gradients, see Section~\ref{sec:appendix:numerical:settings_preliminaries:neurons}.The results in the table illustrate that incorporating reset has the potential to stabilize learning and to improve the performance.

\subsection{Low-bit resolution in neuron communications}
\label{sec:results:activation_functions_comparisons}

We now consider Table \ref{tab:All_Datasets:model_vs_ActFnc} from the perspective of the impact of quantization of the neuron output, i.e. the amplitude resolution in the output of activation function, hence the communication between the neurons.  We recall that {\it{non-signed spikes}} and {\it{signed spikes}} correspond to $1$ and $2$ bit representation in each neuron output whereas in {\it{GELU}}, the output is continuous-valued.  Here, the continuous-valued {\it{GELU}} is included for benchmarking purposes.

% spikes: signed vs nonsigned 
Focusing on the mean accuracy values in our models, we observe that a performance improvement is observed when   {\it{signed}} spikes are used instead of {\it{non-signed}} spikes for all cases. Nevertheless, for many scenarios, the difference  in the accuracy is smaller than  $2\%$, for example MSWC under \stablereset, sMNIST under \stablereset, \stablenoreset, \unstablereset. Hence, the extra bit provided by the {\it{signed}} spiking has not been enough to provide a large meaningful performance improvement in these cases.

% spikes vs cont. valued FOR STABLE+RESET
For \stablereset, the relative performance of {\it{non-signed}} and {\it{signed}} spiking models compared to that of {\it{GELU}} models  depends on the dataset. For MSWC and sMNIST,  the performance is comparable in many cases demonstrating that even with 1-bit quantization (as in the {\it{non-signed}} case), competitive results can still be achieved. 
For DVS-Gesture, the gap between the best performing models under {\it{spiking}}  activation functions and {\it{GELU}} is slightly larger, i.e., $95.4 \%$ and $97.1 \%$, respectively.

% unstable A without reset - GELU
For \unstablenoreset, {\it{GELU}} still performs reasonably well compared to the low-bit spiking neuron models. This suggests that the  communication between neurons, i.e.,  the information transfer between neurons,  may be used to compensate the negative effect of unstable dynamics. In particular, the continuous-valued activation function of {\it{GELU}} allows more information to flow between neurons, which possibly compensates for instability and enables the model to learn relatively effectively.

\subsection{Training stability and convergence}
\label{sec:results:Training_over_epochs}

\begin{figure}
    \centering
    
    \begin{subfigure}[b]{0.49\textwidth}
        \centering
        \includegraphics[width=\textwidth]{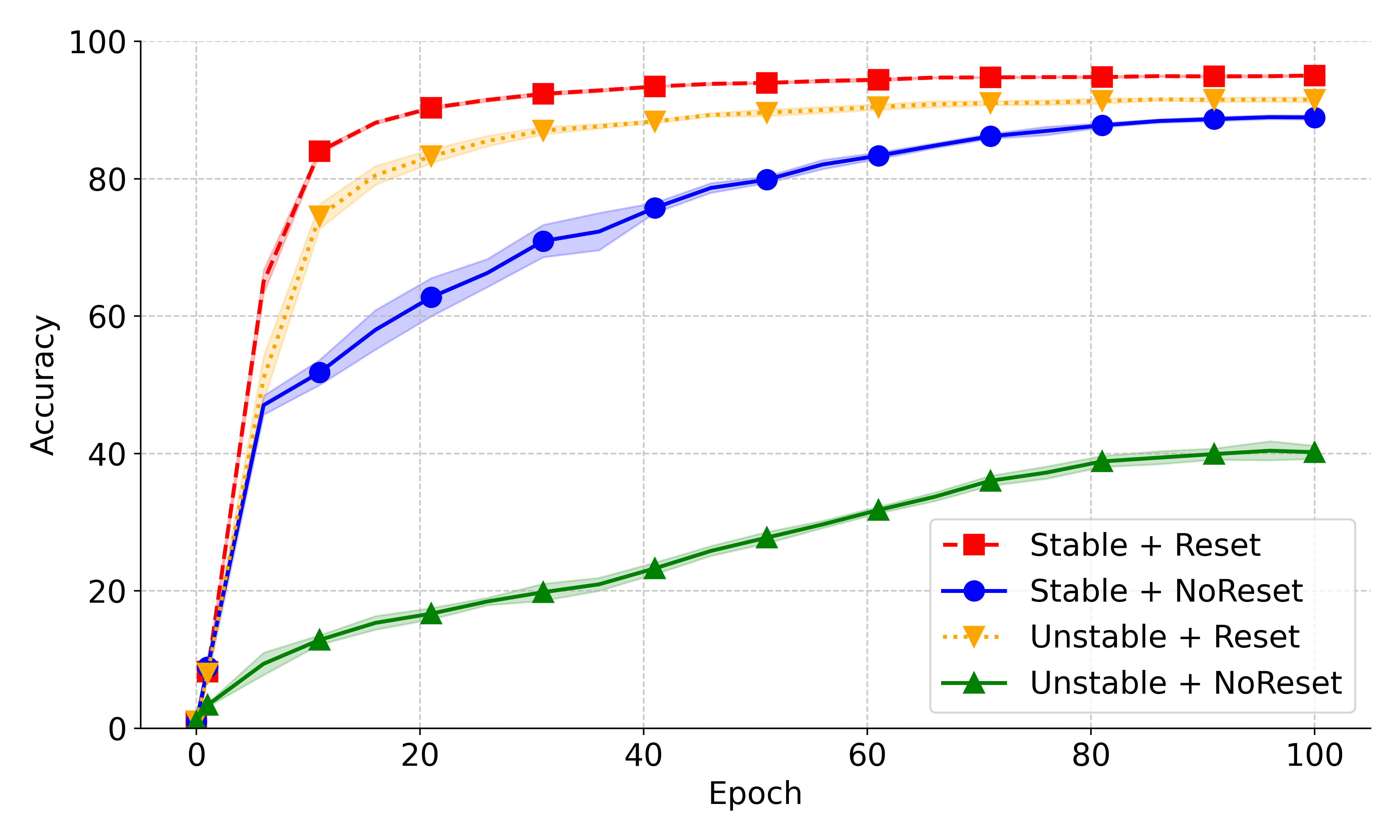}
        \caption{MSWC dataset}
        \label{fig:accuracy_over_epochs:mswc}
    \end{subfigure}
    \hfill
    \begin{subfigure}[b]{0.49\textwidth}
        \centering
        \includegraphics[width=\textwidth]{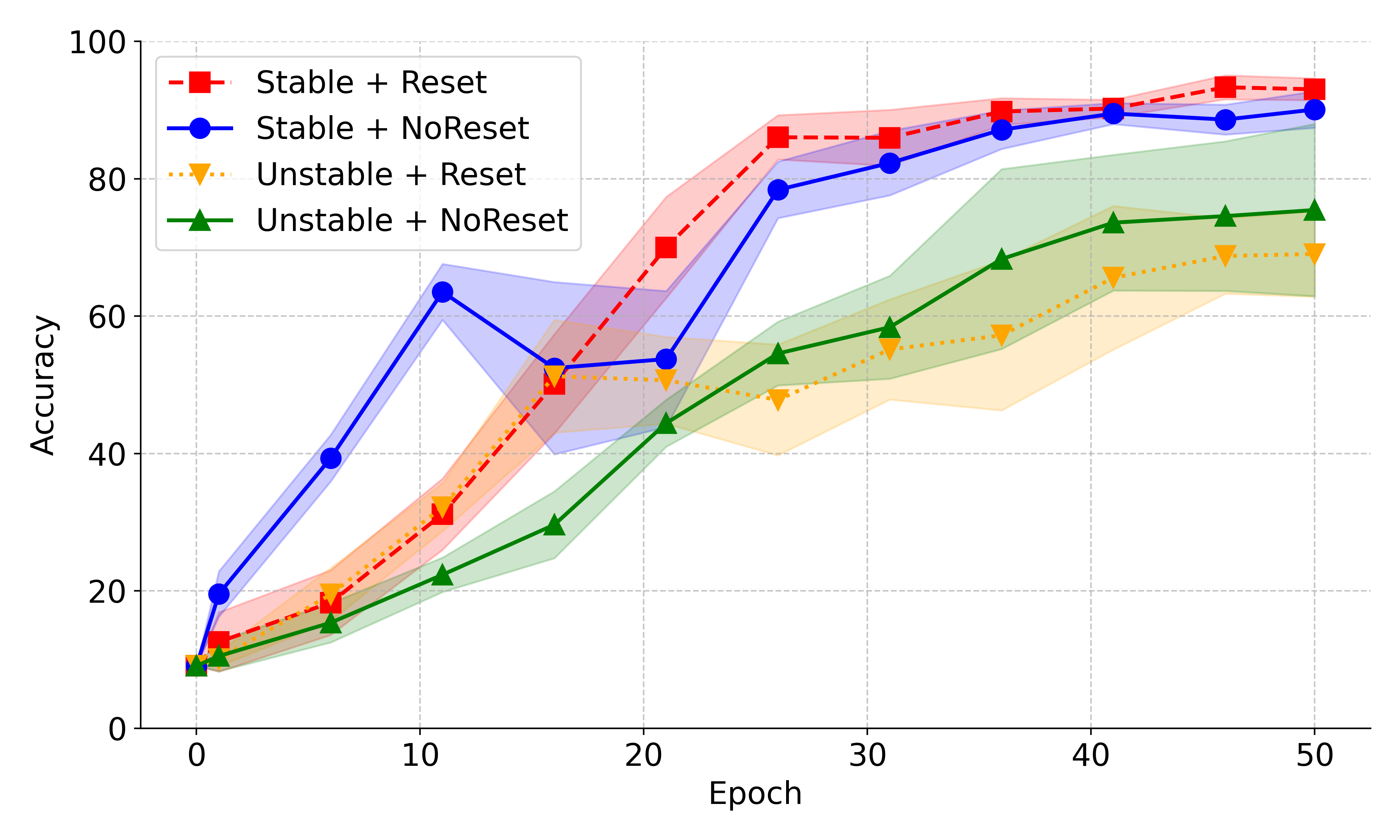}
        \caption{DVS-Gesture dataset}
        \label{fig:accuracy_over_epochs:dvs_gesture}
    \end{subfigure}
    \hfill
    \begin{subfigure}[b]{0.49\textwidth}
        \centering
        \includegraphics[width=\textwidth]{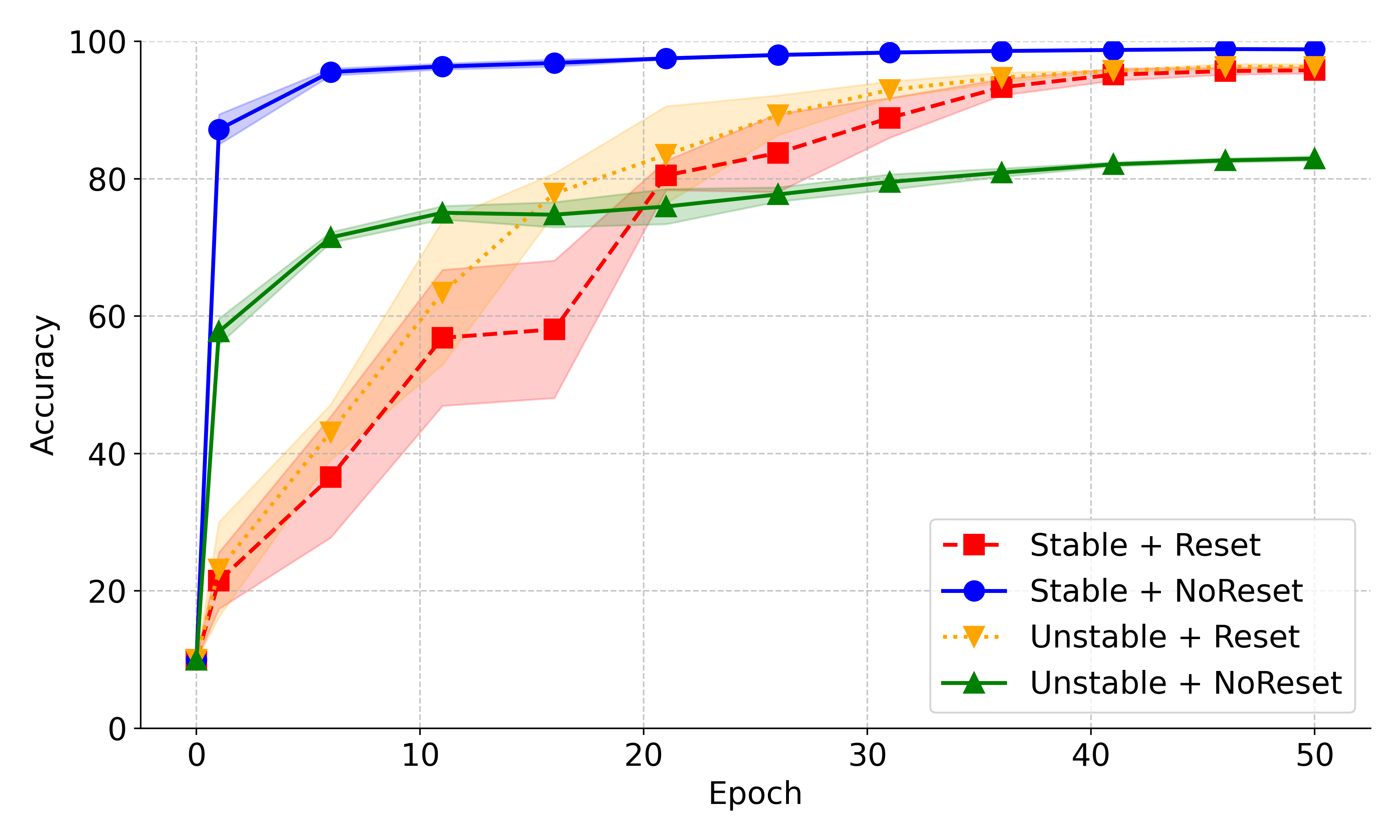}
        \caption{sMNIST dataset}
        \label{fig:accuracy_over_epochs:smnist}
    \end{subfigure}
    
    \caption{{Test classification accuracy over training epochs for {\it{non-signed spikes}}.}}
    \label{fig:accuracy_over_epochs}
\end{figure}

% Intro
We now explore the impact of reset mechanism  and dynamics on the training behavior of the models. Specifically we look at the  convergence and smoothness of accuracy curves in Figure~\ref{fig:accuracy_over_epochs}. In Figure~\ref{fig:accuracy_over_epochs},  test classification accuracy is plotted against the number of training epochs, and the standard deviations across runs are indicated with shaded areas.

% Stable - Reset vs No-Reset
First, we focus on the {\stable} dynamics, and compare the {\stablereset} and {\stablenoreset} models. 
For the MSWC dataset, the model with reset converges faster. 
For the DVS-Gesture dataset,  the model with no reset achieves higher accuracy values earlier during training but then the accuracy values drop to lower values while training continues. Eventually, the model with reset surpasses the model without reset.
For the sMNIST dataset, model without reset  converges faster. 
These results suggest that the impact of the reset mechanism on the convergence during training depends on the characteristics of data. 

% Unstable - Reset vs No-Reset
We now focus on the {\unstable} dynamics, and compare the {\unstablereset} and {\unstablenoreset} models.  For both MSWC and sMNIST, we observe a substantial gap in favor of the model with reset  in terms of the accuracy values the curves  converge to. 
For DVS-Gesture, the models with reset and without reset perform within each other’s standard deviation range. Similar to {\stable} case, the rate of convergence depends on the dataset with no clear trend disfavoring the models with reset.

Across all datasets and for both {\stable}  and {\unstable} models, the accuracy curves in general monotonically increases under reset. The only clearly non-monotonic curve is that of {\stablenoreset} on DVS-Gesture.
We now consider the  initialization dependence, i.e.  the standard deviation ranges across different runs during training.  For the MSWC dataset, there is no significant difference between models with reset and without reset. 
For the DVS-Gesture dataset all models have the same level of variability for a given type of dynamics, i.e. {\stable} or {\unstable}. 
For the sMNIST dataset, models with reset have a higher variability.
Hence, in terms of variation across runs, the presence of reset does not lead to a consistent increase in variability compared to corresponding models without reset.

In conclusion, although our separate reset mechanism introduces an additional non-linear discontinuous data processing block into the spiking neuron, our experiments do not provide any strong evidence that the inclusion of our reset mechanism significantly deteriorates the  convergence speed or training stability. The impact appears to be data-dependent, suggesting that the proposed reset mechanism does not inherently make training more difficult or unstable.

\subsection{Accuracy over time}
\label{sec:results:Training_over_time}

\begin{figure}
    \centering
    
    \begin{subfigure}[b]{0.49\textwidth}
        \centering
        \includegraphics[width=\textwidth]{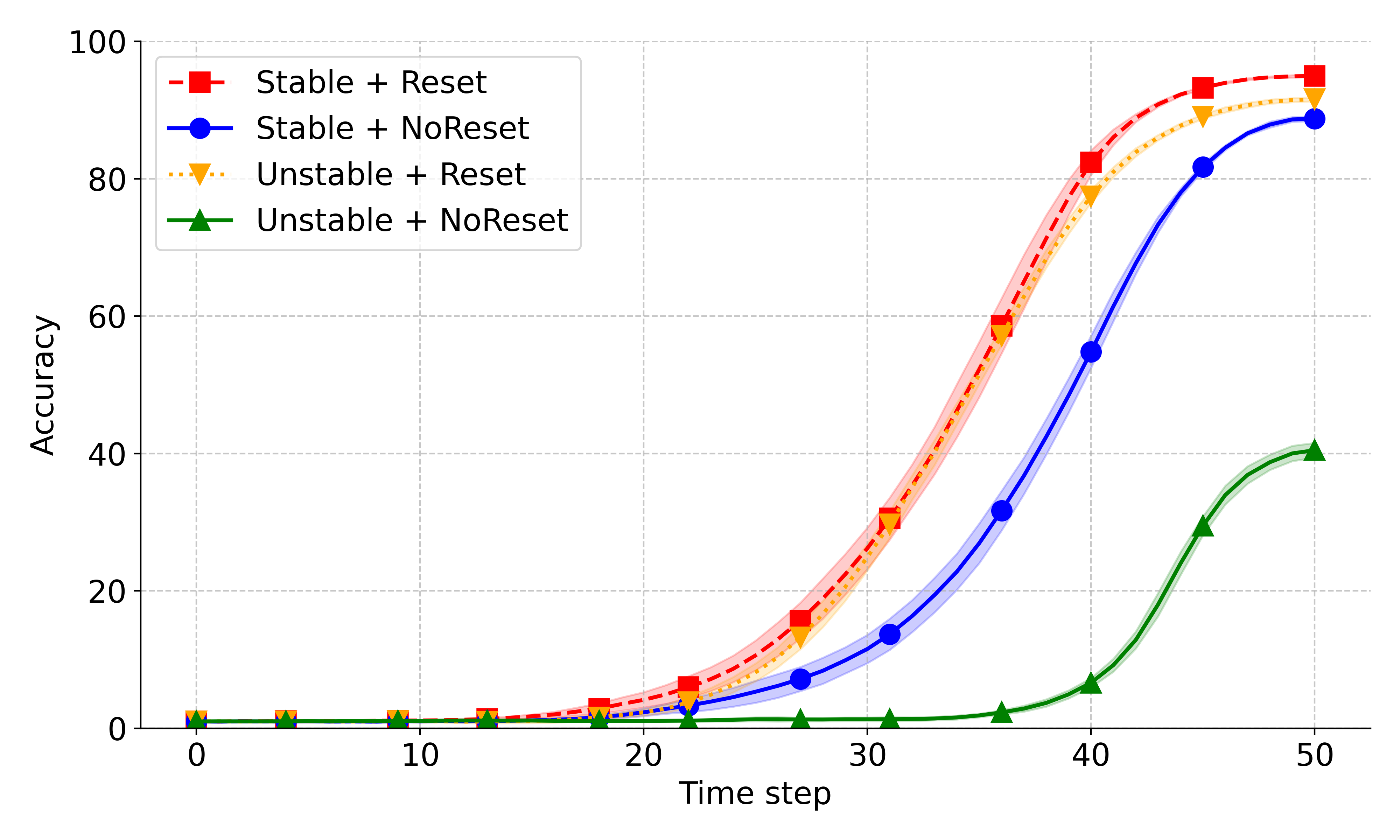}
        \caption{MSWC dataset}
        \label{fig:accuracy_over_time:mswc}
    \end{subfigure}
    \hfill
    \begin{subfigure}[b]{0.49\textwidth}
        \centering
        \includegraphics[width=\textwidth]{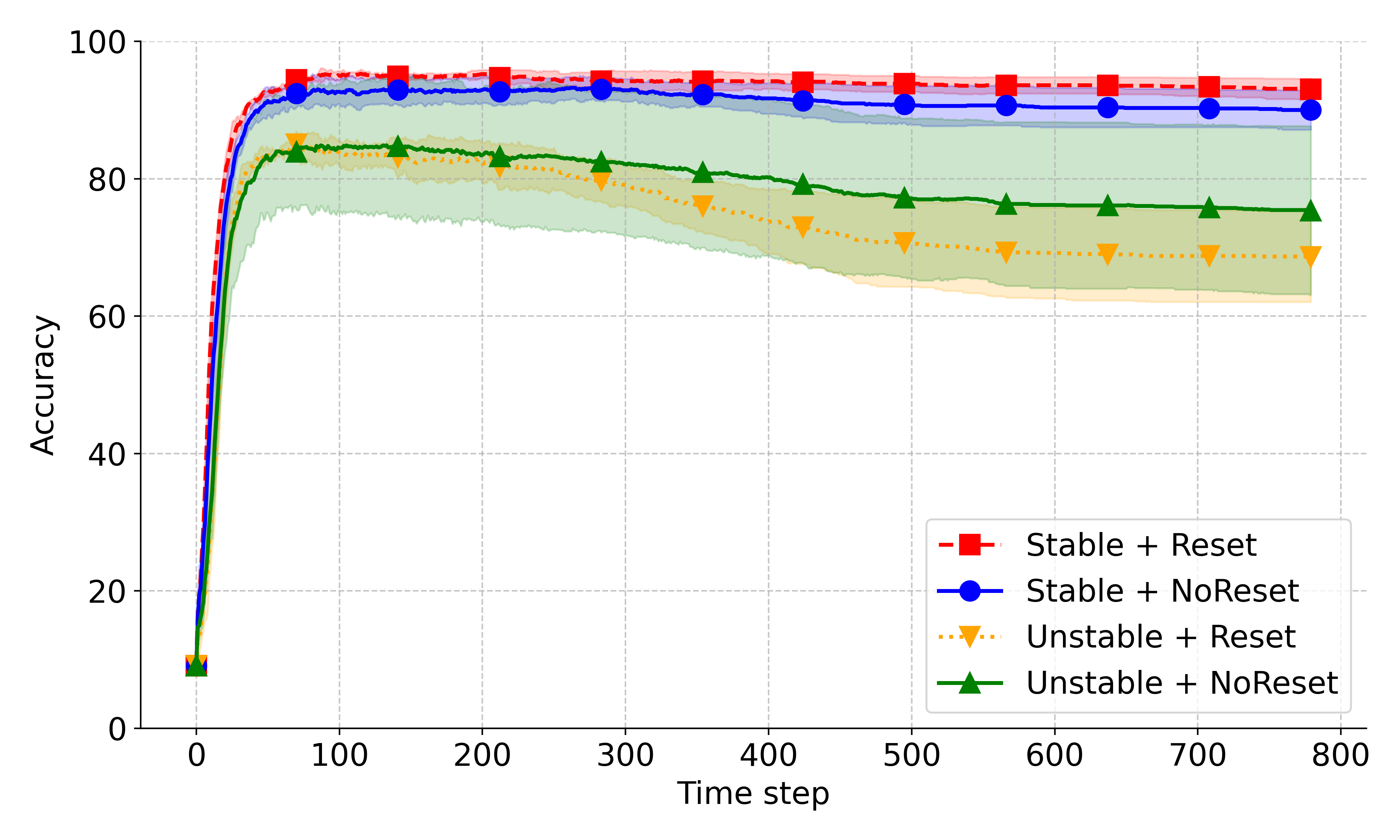}
        \caption{DVS-Gesture dataset}
        \label{fig:accuracy_over_time:dvs_gesture}
    \end{subfigure}
    \hfill
    \begin{subfigure}[b]{0.49\textwidth}
        \centering
        \includegraphics[width=\textwidth]{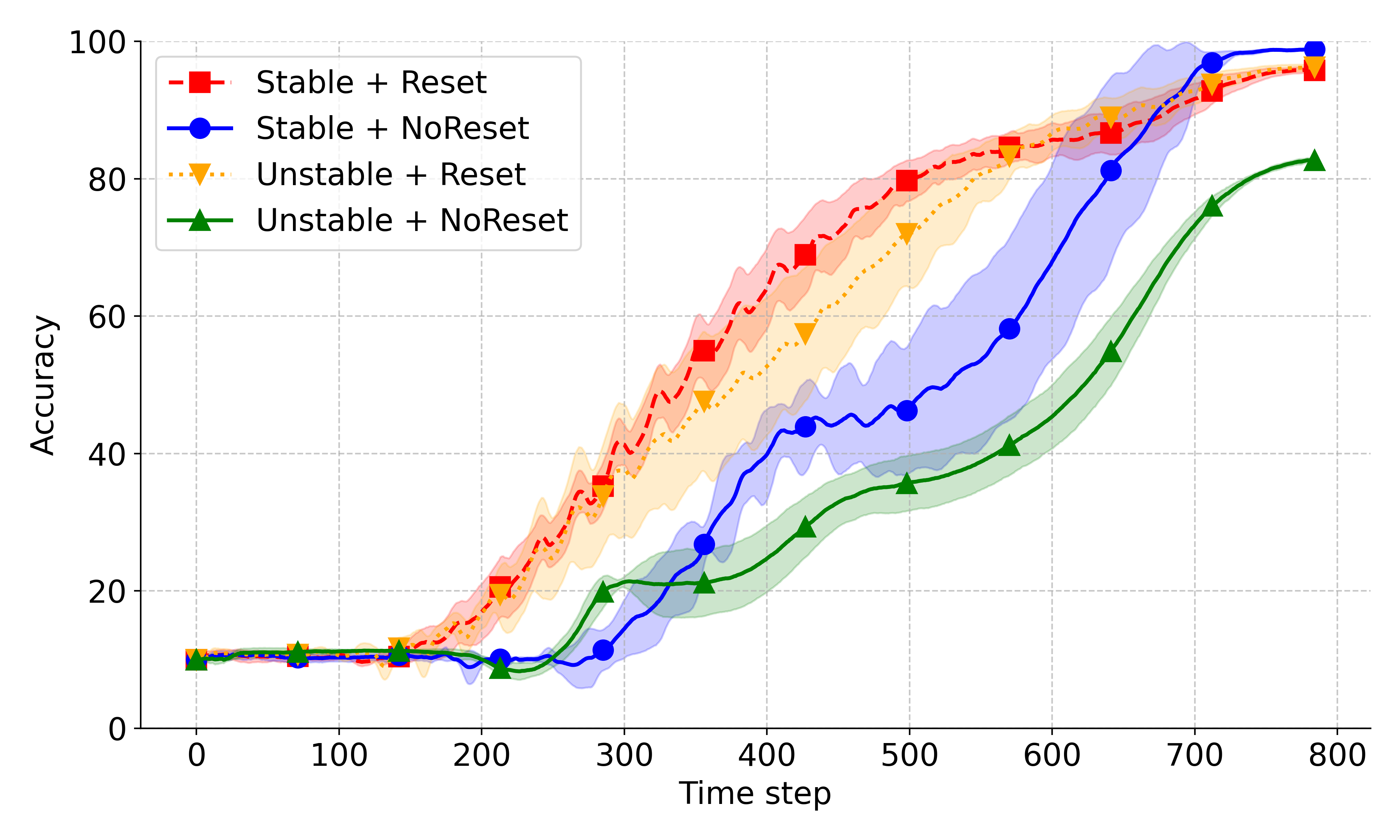}
        \caption{sMNIST dataset}
        \label{fig:accuracy_over_time:smnist}
    \end{subfigure}

    \caption{{Test classification accuracy over time-steps of the input sequences for {\it{non-signed spikes}}.}}
    \label{fig:accuracy_over_time}
\end{figure}

% Intro
In this section, we explore the impact of reset mechanism and dynamics on the accuracy over time. Specifically we look at how the test classification accuracy changes as the models see a larger part of the input sequences. We focus on Figure~\ref{fig:accuracy_over_time} where the test accuracy is plotted against time steps. 

First, we focus on the {\stable} dynamics, and compare the {\stablereset} and {\stablenoreset} models. We observe that for both the MSWC and sMNIST dataset, models with reset achieves reasonable accuracy values faster. Specifically, if we consider the time step at which the models reach $80 \%$ accuracy for {\reset} vs. {\noreset}, we have $t=40$ vs. $t=45$ for MSWC, and $t=500$ vs. $t=640$ for sMNIST. This demonstrates a clear benefit of including the reset mechanism for these datasets if quick decisions at these accuracy levels are desired. For the DVS-Gesture dataset, the performance of {\reset} and {\noreset} configurations is comparable, showing no significant difference. 

Second, we focus on the {\unstable} dynamics, and compare the {\unstablereset} and {\unstablenoreset} models. For MSWC and sMNIST, we observe a substantial accuracy gap in favor of the reset configuration at the end of time sequence. Even setting aside this final  accuracy gap, we observe that models with reset consistently begins improving earlier compared to  the models without reset. 
For DVS-Gesture, the difference in the behavior of the models with and without reset are within the standard deviation ranges of each other.

In conclusion, these experiments illustrate that including the reset can either significantly improve the time to reach a reasonable level of accuracy, or does not  affect it negatively.

\subsection{Spike rate}
\label{sec:results:spike_rate}

\begin{table*}[]
    \centering
    \caption{{Spike rate per hidden layer, rounded to $2$ decimal point. }}
    \label{tab:All_Datasets:spike_rate}
    \newcolumntype{l}{>{\centering\arraybackslash}p{1.2cm}}
    \newcolumntype{k}{>{\centering\arraybackslash}p{0.9cm}}
    \newcolumntype{m}{>{\centering\arraybackslash}p{1.5cm}}

    \begin{subtable}[h]{1\linewidth} 
    \centering
    \vspace{3pt}
    \caption{MSWC dataset}
    \label{tab:mswc:spike_rate}
    \begin{tabular}{lkmmmm}
    \hline
        Dynamics & Reset
               & \multicolumn{2}{c}{{\it{Non-signed spikes}}}
               & \multicolumn{2}{c}{{\it{Signed spikes}}}\\
    \hline
                && layer $1$ & layer $2$
                 & layer $1$ & layer $2$\\
    \hline
    Stable & Yes &0.42 $\pm$ 0.15 & 0.40 $\pm$ 0.16 & 0.69 $\pm$ 0.15 & 0.78 $\pm$ 0.16 \\
    Stable & No &0.46 $\pm$ 0.28 & 0.40 $\pm$ 0.31 & 0.98 $\pm$ 0.03 & 0.98 $\pm$ 0.04 \\
    Unstable & Yes&0.49 $\pm$ 0.15 & 0.37 $\pm$ 0.16 & 0.70 $\pm$ 0.15 & 0.75 $\pm$ 0.16 \\
    Unstable & No &0.50 $\pm$ 0.21 & 0.49 $\pm$ 0.20 & 0.98 $\pm$ 0.03 & 0.99 $\pm$ 0.01 \\
    \hline
    \end{tabular}
    \end{subtable}
    
    \begin{subtable}[h]{1\linewidth} 
    \centering
    \vspace{3pt}
    \caption{DVS-Gesture dataset}
    \label{tab:dvs:spike_rate}
    \begin{tabular}{lkmmmmmm}
    \hline
        Dynamics & Reset
               & \multicolumn{3}{c}{{\it{Non-signed spikes}}}
               & \multicolumn{3}{c}{{\it{Signed spikes}}}\\
    \hline
                && layer $1$ & layer $2$ & layer $3$
                 & layer $1$ & layer $2$ & layer $3$\\
    \hline
    Stable & Yes  &0.50 $\pm$ 0.27 & 0.50 $\pm$ 0.26 & 0.50 $\pm$ 0.28 & 0.80 $\pm$ 0.20 & 0.92 $\pm$ 0.14 & 0.98 $\pm$ 0.09 \\
    Stable & No &0.50 $\pm$ 0.28 & 0.50 $\pm$ 0.26 & 0.50 $\pm$ 0.28 & 0.74 $\pm$ 0.23 & 0.85 $\pm$ 0.19 & 0.95 $\pm$ 0.12 \\
    Unstable & Yes &0.50 $\pm$ 0.28 & 0.50 $\pm$ 0.27 & 0.50 $\pm$ 0.28 & 0.66 $\pm$ 0.25 & 0.81 $\pm$ 0.22 & 0.92 $\pm$ 0.15 \\
    Unstable & No &0.50 $\pm$ 0.27 & 0.50 $\pm$ 0.28 & 0.50 $\pm$ 0.30 & 0.65 $\pm$ 0.25 & 0.80 $\pm$ 0.22 & 0.90 $\pm$ 0.17 \\
    \hline
    \end{tabular}
    \end{subtable}

    \begin{subtable}[h]{1\linewidth} 
    \centering
    \vspace{3pt}
    \caption{sMNIST dataset}
    \label{tab:smnist:spike_rate}
    \begin{tabular}{lkmmmm}
    \hline
        Dynamics & Reset
               & \multicolumn{2}{c}{{\it{Non-signed spikes}}}
               & \multicolumn{2}{c}{{\it{Signed spikes}}}\\
    \hline
                && layer $1$ & layer $2$
                 & layer $1$ & layer $2$\\
    \hline
    Stable & Yes &0.46 $\pm$ 0.30 & 0.49 $\pm$ 0.22 & 0.71 $\pm$ 0.30 & 0.77 $\pm$ 0.23 \\
    Stable & No &0.50 $\pm$ 0.12 & 0.50 $\pm$ 0.09 & 0.97 $\pm$ 0.04 & 0.98 $\pm$ 0.03 \\
    Unstable & Yes &0.48 $\pm$ 0.24 & 0.49 $\pm$ 0.16 & 0.80 $\pm$ 0.22 & 0.84 $\pm$ 0.17 \\
    Unstable & No &0.50 $\pm$ 0.09 & 0.50 $\pm$ 0.07 & 1.00 $\pm$ 0.00 & 1.00 $\pm$ 0.00 \\
    \hline
    \end{tabular}
    \end{subtable}
\end{table*}

% Calculation
We now investigate the spike rate for the models with spiking activation function.
% Each spiking neuron emits spikes over time,
At each time step, each spiking neuron either provides no output,   or emits a spike, either always with a value of $1$, as in {\it{non-signed spikes}}, or with values of $1$ or $-1$, as in {\it{signed spikes}}. In this experiment, we focus on spike activity regardless of its value and calculate the spike rate over time for all neurons in a hidden layer and calculate its mean and standard deviation. The values in Table \ref{tab:All_Datasets:spike_rate} represent the mean of these values for our models from Table \ref{tab:All_Datasets:model_vs_ActFnc}.

% Observation
In particular, we observe two cases where the mean spike rate is significantly lower when reset is used, compared to configurations without reset: for MSWC, {\it{signed spikes}} and {\it{Stable}} configuration, where including the reset reduced the spike rate from $0.98$ to $0.69$ in layer $1$, and $0.99$ to $0.75$ in layer $2$. In all other cases examined, no significant difference is observed i.e., the mean values are within one standard deviation range of each other. 
We note that no explicit regularization on spike rate was imposed during training. 
Nevertheless, the results suggest that, across the studied datasets, the proposed reset mechanism alone either reduces the spike rate or leaves it unchanged.

\subsection{Multiple output channels in spiking neurons}
\label{sec:results:Drop_Output_channels}
We now investigate whether the output channels of neurons are equally informative or exhibit redundancy.

To investigate this aspect, we perform a series of experiments where a number of output channels of each neuron are dropped, i.e. the output of the neuron for that output channel is set to $0$ at all time steps. The channels that are dropped are selected based on their indices using either the ``First'' or ``Last'' indices. For instance, if $2$ channels are dropped out of $\nout$ channels, then the channels with indices $1$ and $2$ are dropped under ``First'', while the channels with indices $\nout-1$ and $\nout$ are dropped under ``Last''. 
We use the pretrained models of Table~\ref{tab:All_Datasets:model_vs_ActFnc} for {\stablereset} with {\it{non-signed}} spikes for each dataset.
We present the resulting accuracy values of this experiment in Table~\ref{tab:All_Datasets:Drop_Output_channels}.

Across all datasets, we observe that removing a single output channel per neuron leads to a noticeable drop in accuracy, which becomes more pronounced when two channels are removed. This suggests that each output channel captures complementary features that cannot be fully recovered by the others. We now compare the results across different rows, i.e., when varying which specific output channel(s) are removed based on their index.  Considering the standard deviations, we observe that the performance remains within the same accuracy interval in each case. This suggests that all output channels are similarly important and contribute comparably to the network’s overall performance.

\begin{table}[]
    \centering
    \caption{{Classification accuracy (\%) under drop-out of output channels of neurons for the models configuration {\stablereset} under {\it{non-signed spikes}}}}.
    \label{tab:All_Datasets:Drop_Output_channels}
    \newcolumntype{l}{>{\centering\arraybackslash}p{2.5cm}}
    \newcolumntype{m}{>{\centering\arraybackslash}p{2.1cm}}

    %----------------------------------------------------
    \begin{subtable}[h]{1\linewidth} 
    \centering
    \vspace{3pt}
    \caption{MSWC dataset. Recall $\nout=4$,  and the baseline from Table \ref{tab:mswc:model_vs_ActFnc}, which uses all output channels of neurons is $95.0 \pm 0.1\%$.}
    \label{tab:mswc:Drop_Output_channels}
    \begin{tabular}{lmm}
    \hline
    \multirow{1}{*}{Which \textbackslash How many} 
        & drop 1 channel  
        & drop 2 channels \\
    \hline
    First  &48.0 $\pm$ 9.5 \% & 4.3 $\pm$ 1.5 \% \\
    Last &46.7 $\pm$ 4.2 \% & 3.8 $\pm$ 2.0 \% \\
    \hline
    \end{tabular}
    \end{subtable}
    %----------------------------------------------------
    
    %----------------------------------------------------
    \begin{subtable}[h]{1\linewidth} 
    \centering
    \vspace{3pt}
    \caption{DVS-Gesture dataset. Recall $\nout=24$, and the baseline from Table \ref{tab:dvs:model_vs_ActFnc}, which uses all output channels of neurons is $93.0 \pm 1.8\%$.}
    \label{tab:dvs:Drop_Output_channels}
    \begin{tabular}{lmm}
    \hline
    \multirow{1}{*}{Which \textbackslash How many} 
        & drop 1 channel  
        & drop 2 channels \\
    \hline
    First  & 79.0 $\pm$ 7.4 \% & 65.9 $\pm$ 5.3 \% \\
    Last & 70.9 $\pm$ 7.2 \% & 63.5 $\pm$ 14.1 \% \\
    \hline
    \end{tabular}
    \end{subtable}
    %----------------------------------------------------

    %----------------------------------------------------
    \begin{subtable}[h]{1\linewidth} 
    \centering
    \vspace{3pt}
    \caption{sMNIST dataset. Recall $\nout=8$,  and the baseline from Table~\ref{tab:smnist:model_vs_ActFnc}, which uses all output channels of neurons is $95.8 \pm 0.5\%$.}
    \label{tab:smnist:Drop_Output_channels}
    \begin{tabular}{lmm}
    \hline
    \multirow{1}{*}{Which \textbackslash How many} 
        & drop 1 channel  
        & drop 2 channels \\
    \hline
    First  &40.7 $\pm$ 11.8 \% & 19.6 $\pm$ \, 9.8 \% \\
    Last &40.4 $\pm$ 17.5 \% & 26.4 $\pm$ 13.8 \% \\
    \hline
    \end{tabular}
    \end{subtable}
    %----------------------------------------------------
\end{table}

\subsection{Impact of network architecture}
\label{sec:results:mswc-H_vs_N_vs_Nout}

% MSWC table with varying H, N and N-out
\begin{table*}[]
    \centering
    \caption{Classification accuracy ($\%$) on the MSWC dataset for different models by varying the neuron dynamics and dimensions. 
    The best accuracy over all models is in bold.}
    \label{tab:mswc:models_vs_nout_vs_n}
    \newcolumntype{l}{>{\centering\arraybackslash}p{1.7cm}}
    \newcolumntype{k}{>{\centering\arraybackslash}p{1cm}}
    \newcolumntype{m}{>{\centering\arraybackslash}p{3cm}}
    %----------------------------------------------------
    \begin{subtable}[h]{1\linewidth}
    \centering
    \vspace{3pt}
    \caption{Models have $\Hneurons=1024$ neurons per hidden layer, each with $\Nstate=2$ state variables. 
    Here, $\nout=1$ satisfies $\Hneurons \times \nout=1024$.}
    \begin{tabular}{lkmmmm}
    \hline
       Dynamics 
       & Reset
       & $\nout=1$
       & $\nout=4$
       & $\nout=8$ 
       & $\nout=16$ \\
    \hline
    Stable & Yes&95.2  $\pm$  0.2 \% & 95.7  $\pm$  0.1 \% & \textbf{95.8  $\pm$  0.1} \% & 95.7  $\pm$  0.1 \% \\
    Stable & No &93.8  $\pm$  0.1 \% & 95.3  $\pm$  0.1 \% & 95.6  $\pm$  0.0 \% & \textbf{95.8  $\pm$  0.1} \% \\
    Unstable & Yes &91.6  $\pm$  0.2 \% & 93.5  $\pm$  0.1 \% & 93.3  $\pm$  0.4 \% & 93.9  $\pm$  0.2 \% \\
    Unstable & No  &79.1  $\pm$  0.6 \% & 84.4  $\pm$  0.2 \% & 86.8  $\pm$  0.3 \% & 88.4  $\pm$  0.4 \% \\
    \hline
    \end{tabular}
    \end{subtable}
    %----------------------------------------------------
    \begin{subtable}[h]{1\linewidth}
    \centering
    \vspace{3pt}
    \caption{Models have $\Hneurons=256$ neurons per hidden layer, each with $\Nstate=8$ state variables. 
    Here, $\nout=4$ satisfies $\Hneurons  \times  \nout=1024$.}
    \begin{tabular}{lkmmmm}
    \hline
       Dynamics 
       & Reset
       & $\nout=1$
       & $\nout=4$
       & $\nout=8$ 
       & $\nout=16$ \\
    \hline
    Stable & Yes &89.3  $\pm$  0.7 \% & 94.9  $\pm$  0.1  \% &  95.2  $\pm$  0.1 \% & 95.4 $\pm$ 0.1 \% \\
    Stable & No &78.0  $\pm$  0.6 \% & 89.0  $\pm$  0.2 \% & 91.0  $\pm$  0.1 \% & 92.5  $\pm$  0.3 \% \\
    Unstable & Yes &81.0  $\pm$  1.1 \% & 91.6  $\pm$  0.3 \% & 94.4  $\pm$  0.3 \% & 94.6  $\pm$  0.4 \% \\
    Unstable & No &20.7  $\pm$  1.1 \% & 40.0  $\pm$  0.7 \% & 49.9  $\pm$  0.9 \% & 59.2  $\pm$  0.5 \% \\
    \hline
    \end{tabular}
    \end{subtable}
    %----------------------------------------------------
    \begin{subtable}[h]{1\linewidth}
    \centering
    \vspace{3pt}
    \caption{Models have $\Hneurons=128$ neurons per hidden layer, each with $\Nstate=16$ state variables. 
    Here, $\nout=8$ satisfies $\Hneurons \times  \nout=1024$.}
    \begin{tabular}{lkmmmm}
    \hline
       Dynamics 
       & Reset
       & $\nout=1$
       & $\nout=4$
       & $\nout=8$ 
       & $\nout=16$ \\
    \hline
    Stable & Yes &  83.3  $\pm$  1.0 \% & 93.5  $\pm$  0.3 \% & 94.5  $\pm$  0.2 \% & 95.0  $\pm$  0.1 \% \\
    Stable & No&63.4  $\pm$  2.3 \% & 85.0  $\pm$  0.2 \% & 88.3  $\pm$  0.3 \% & 90.4  $\pm$  0.2 \% \\
    Unstable & Yes &69.7  $\pm$  1.8 \% & 86.8  $\pm$  1.0 \% & 92.4  $\pm$  1.2 \% & 94.4  $\pm$  0.3 \% \\
    Unstable & No&\,8.3  $\pm$  0.4 \% & 19.3  $\pm$  0.8 \% & 26.6  $\pm$  0.6 \% & 35.1  $\pm$  0.7 \% \\
    \hline
    \end{tabular}
    \end{subtable}
    %----------------------------------------------------
    \begin{subtable}[h]{1\linewidth}
    \centering
    \vspace{3pt}
    \caption{Models have $\Hneurons=64$ neurons per hidden layer, each with $\Nstate=32$ state variables. 
    Here, $\nout=16$ satisfies $\Hneurons \times  \nout=1024$.}
    \begin{tabular}{lkmmmm}
    \hline
       Dynamics 
       & Reset
       & $\nout=1$
       & $\nout=4$
       & $\nout=8$ 
       & $\nout=16$ \\
    \hline
    Stable & Yes & 70.1  $\pm$  2.9 \% & 90.3  $\pm$  0.3 \% & 92.8  $\pm$  0.3 \% & 94.0 $\pm$ 0.2 \% \\
    Stable & No &42.7  $\pm$  1.4 \% & 77.5  $\pm$  1.0 \% & 84.2 $\pm$  0.6 \% & 87.4  $\pm$  0.3 \% \\
    Unstable & Yes &53.3  $\pm$  2.4 \% & 77.8  $\pm$  1.8 \% & 85.4  $\pm$  1.0 \% & 91.8  $\pm$  1.3 \% \\
    Unstable & No & \, 3.7  $\pm$  0.4 \% & \, 8.2  $\pm$  0.3 \% & 13.8  $\pm$  0.9 \% & 19.3  $\pm$  0.7 \% \\
    \hline
    \end{tabular}
    \end{subtable}
\end{table*}

We now investigate the effect of network architecture. 
For illustration,  we focus on the MSWC dataset and provide the results in Table \ref{tab:mswc:models_vs_nout_vs_n} for {\it{non-signed spikes}}.

The architectures in Table~ \ref{tab:mswc:models_vs_nout_vs_n}  are designed as follows: 
For fair comparison in terms of layer width and network depth, we consider the SNN  architectures in \cite{yik2024neurobench, karilanova2024zeroshottemporalresolutiondomain} for MSWC dataset as a starting point. In particular, both works use SNN models of $2$ hidden layers with $\Hneurons=1024$ adLIF neurons per hidden layer, where each neuron has two state variables, i.e.,  $\Nstate=2$, and has a single output, i.e., $\nout=1$. Hence, in all experiments in Table~ \ref{tab:mswc:models_vs_nout_vs_n},   we use $2$ hidden layers and explore the trade-offs between different layer architectures in terms number of neurons $\Hneurons$, the number of neuron state dimension $\Nstate$, and number of output channels per neuron $\nout$. We consider two design rules as reference points:
\begin{itemize}
    \item Fixed latent space dimension: number of total state variables in  a hidden layer is fixed as $\Hneurons \times \Nstate=2048$
    \item  Fixed number of outgoing synaptic connections: number of outgoing synaptic connections from a hidden layer is fixed as  $\Hneurons \times \nout=1024$
\end{itemize}

In Table~\ref{tab:mswc:models_vs_nout_vs_n},  every model  has a constant number of state variables in a layer, $\Hneurons \times \Nstate=2048$. Some of the  models  further satisfy $\nout \times \Hneurons=1024$, as indicated in the captions.

Table~\ref{tab:mswc:models_vs_nout_vs_n} illustrates that there are multiple network architectures that achieve comparable performance under the proposed reset mechanism. 
For instance, all the network architectures with $h=1024$, $n=2$ under {\stable} with the proposed reset obtain above $95\%$ average accuracy. Table~\ref{tab:mswc:models_vs_nout_vs_n} also illustrates that the proposed reset mechanism may provide performance improvements over a wide range of network architectures. 
For {\stable} case, inclusion of reset results in either a change in the performance within the standard deviation range, or an improvement of up to $\approx 38$\%  where the magnitude of improvement is typically larger with lower $\nout$ and lower $h$. For the {\unstable} case, it is crucial to use reset to obtain  performance levels comparable with the {\stable} case. 
Comparing {\unstablereset} with \unstablenoreset, we observe that inclusion of reset introduces a performance improvement in the range $\approx[5\%, 60\%]$, where again,  the magnitude of improvement is typically larger with lower $\nout$ and lower $h$. 
\section{Discussions and Conclusions}
\label{sec:discussions_and_conclusions}

This work proposes a multiple-output, stateful spiking neuron model with a general linear state transition dynamics and non-linear  reset-based feedback mechanism. Inspired by neuromorphic computing principles, the proposed model incorporates spike-based communication, threshold mechanisms, and state discharging based on its output.

Our formulation generalizes the reset mechanisms in SNN literature by providing a clear formulation that distinguishes  the spiking function, the reset condition, and  the reset action. Our results illustrate that  competitive performance can be obtained using the proposed neuron model.

An important aspect of the proposed reset mechanism is its role in the stability of the neuron,  and hence stability of learning in the neural network. In contrast to the existing works that enforce stable linear SSM transition, our proposed neuron model allows instability in the linear part of  the state transition. In particular, our results illustrate that  for models with  unstable linear dynamics and low-bit information processing, i.e., activation functions whose outputs are quantized with a low number of bits, reset mechanism may make learning possible even when the trainable network parameters diverge and the network fails to learn without the reset mechanism. Our results showed that across different datasets and for a wide range of  network architectures reset   contributed to achieving results comparable with the benchmark of continuous-valued information processing. Hence, our framework provides a promising starting point for investigation of a richer family of SSM models in neural networks.

We now discuss the implementation of the proposed neuron models on neuromorphic hardware.
LIF type of reset mechanism has been implemented on multiple neuromorphic chips such as Loihi \cite{davies_advancing_2021}, TrueNorth \cite{TrueNorth_2014}. By representing complex-valued matrices with real-valued equivalents, feasibility of deploying models with complex-valued diagonal state-space transition matrices on Loihi has been recently demonstrated \cite{meyer2024diagonalstructuredstatespace}. However, whether our proposed model, which combines a reset mechanism that goes beyond the reset mechanism in LIF, and complex-valued state-space models, may be efficiently deployed on neuromorphic hardware remains an open question. Investigation of feasibility of such an implementation,  and the resulting accuracy, computational complexity and energy trade-offs is a key direction for future work.

A fundamental question is to what extent the reset mechanism can prevent divergence in learning under neurons with an unstable dynamics and low-bit activation functions. Quantification of these limits is considered a promising future research direction. Another future research direction is exploring fully learnable reset conditions. This article investigated a reset mechanism where the reset condition is given by a norm condition on the state variable with a learnable effective threshold parameter and the reset action is performed using  linear scaling with a learnable parameter. More general forms of reset conditions and actions, where the parameters of broader family of functions are learned from data, is considered a promising research direction.
\section{Appendix}\label{sec:appendix}

\subsection{Further Details of the  Set-up  for Numerical Results}
\label{sec:appendix:numerical:settings}

\subsubsection{SSM-Neuron Parametrization}\label{sec:appendix:numerical:settings_preliminaries:neurons}

% Initialization of A
 The diagonal matrix $\Lambdabf$ is initialized using the S4D-Lin initialization with bilinear discretization \cite{gu2022parameterization}, where $\Nstate$ different eigenvalues are initialized without explicit repetition of complex conjugates. 

% stable A
For stable $\Abf$, this initialization of $\Lambdabf$ is used with an additional clipping of the modulus of  eigenvalues so that they satisfy $\leq1$ while training in order to ensure stable discrete system/neuron. 
% un-stable A
On the other hand,  for unstable $\Abf$, after the initialization of $\Lambdabf$, we perform additional step by multiplying every second eigenvalues by $1.5$. This results in unstable system/neuron where half of the eigenvalues of $\Abf$ have modulus $>1$. Note due to the delta range in the initialization of $\Lambdabf$ \cite{gu2022parameterization}, modulus of the eigenvalues $\approx 0.9$, and hence multiplying the  eigenvalues with $1.5$ results in modulus of eigenvalues $>1$, hence unstable system/neuron. In unstable dynamics, we do not perform clipping of eigenvalues while training, allowing the eigenvalues of $\Abf$ to settle for possibly unstable values. 

% un-stable A + no reset
For unstable $\Abf$, 
$\|\vbf\|$ possibly increases in unbounded fashion. If no action is taken, this typically leads to very large numbers in the state, and  prevents learning completely.  Hence,  
we clip as $|v_k|<1000$, which prevents exploding of state values while training. 
%

% Rest of the SSM params
Following the S4D-Lin initialization with bilinear discretization \cite{gu2022parameterization}, $\Cbf$ is initialized from the standard normal distribution and $\Bbf$ is fixed as a real-valued matrix of ones. 
We initialize $\CbfBias$ with zeros. Hence, the trainable parameters in the linear part of the SSM are $\Lambdabf$,  $\Cbf$ and $\CbfBias$.

% Complex Output
The output of the state evolution $\ybf$ is complex-valued. We use the transformation $\ybf \rightarrow \Re(\ybf)+\Im(\ybf)$ before applying the  activation function $\ActFnc(\dot)$.

\subsubsection{Training}\label{sec:appendix:numerical:settings_preliminaries:training}
%Hardware
Models are trained on NVIDIA Tesla T4 GPUs with 16GB RAM.
% HPO
Due to computational constraints, limited hyperparameters optimization (HPO) is performed and hyperparameters based on a validation subset are chosen, see Table \ref{tab:hyperparams} for the resulting values. 
%% sMNIST and DVS-Gesture
%
For sMNIST and DVS-Gesture datasets,  HPO was performed for the case of {\stablenoreset} and GELU activation function,  and the chosen hyperparameters are used for all other model cases and activation functions. For the reset-dependent hyperparameters (i.e. initial values of $\RstValue$ and $\RstBias$ and related learning rates), an additional HPO is performed on {\stablereset} and GELU activation function and the chosen hyperparameters are used for all other model cases with reset and activation functions.
%
%% MSWC
For MSWC dataset, HPO  was performed for the cases of (i) {\stablereset}, (ii) {\stablenoreset} and (iii) {\unstablereset}. For each case,  the hyper-parameters evaluated are the base learning rates, weight decay regularization parameters, $\Nstate$ and $\nout$. In addition, for the {\it{Reset}} case the initial value of $\RstValue$ and $\RstBias$ are considered. We do not perform explicit search for LR and WD parameters related to $\RstValue$ and $\RstBias$, but instead use the LR-others and WD-others, respectively. The choices from (iii) are also used for {\unstablenoreset}. For all HPO studies of MSWC, we use {\it{non-signed spikes}} activation function. The same hyper-parameter set is used for all multiple-output dimensions scenarios in Table \ref{tab:mswc:models_vs_nout_vs_n} and all activation functions in Table \ref{tab:mswc:model_vs_ActFnc}.

% Other training info
In all models, ADAMW optimizer and cosine learning rate scheduler are adopted. To prevent vanishing or exploding gradients which might occur while training, we clip the gradients' absolute value to $10^{5}$.

\begin{table}[]
    \centering
    \caption{Hyperparameters used for each dataset.}
    \label{tab:hyperparams}
    \begin{threeparttable}
    \newcolumntype{l}{>{\centering\arraybackslash}p{2.5cm}}
    \newcolumntype{m}{>{\centering\arraybackslash}p{1.6cm}}
    \newcolumntype{k}{>{\centering\arraybackslash}p{1.4cm}}
    %----------------------------------------------------
    \begin{tabular}{lmkk}
    \hline % 32768=(128*128*2)
    Parameter 
        & MSWC
        & DVS-Gesture
        & sMNIST \\
    \hline
    $\cin$ & 20 & 32768 & 1\\
    $\cout$ & 100 & 11 & 10\\
    Nb. Hidden Layers & 2 & 3 & 2\\
    $\Hneurons$ & 256 \tnote{(*)} & 128 & 96 \\
    $\Nstate$  & 8 \tnote{(*)} & 8 & 8 \\
    $\nout$   & 4 \tnote{(*)} & 24 & 8 \\
    LR &See Table~\ref{tab:hyperparams_mswc}&$10^{-4}$&$10^{-3}$ \\
    LR-SSM &See Table~\ref{tab:hyperparams_mswc}&$10^{-2}$&$10^{-4}$ \\
    LR-$\RstValue$ &=LR&$10^{-4}$&$10^{-5}$ \\
    LR-$\RstBias$ &=LR&$10^{-6}$&$10^{-5}$ \\
    WD  &See Table~\ref{tab:hyperparams_mswc}&$0$&$10^{-2}$ \\
    WD-SSM  &$10^{-3}$&$10^{-2}$&$10^{-3}$ \\
    WD-$\RstValue$  & =WD &$10^{-2}$&$0$ \\
    WD-$\RstBias$  & =WD &$10^{-5}$&$0$ \\
    $\RstValue$ initialization & 0.1 & 0.8 & 0.5 \\
    $\RstBias$ initialization  & 0 & 0 & 0 \\
    Nb. Epochs & 100 & 50 & 50 \\
    Batch Size  & 256 & 16 & 128 \\
    Dropout  & 0.4 & 0 & 0.3 \\
    \hline
    \end{tabular}
    %----------------------------------------------------
    \begin{tablenotes}
    \small
        \item[(*)] The given value is used for all MSWC results in Section~\ref{sec:results} except in Table~\ref{tab:mswc:models_vs_nout_vs_n}, where it is explicitly stated otherwise.
    \end{tablenotes}
    \end{threeparttable}
\end{table}

\begin{table}[]
    \centering
    \caption{Additional hyperparameters used for MSWC dataset.}
    \label{tab:hyperparams_mswc}
    \newcolumntype{l}{>{\centering\arraybackslash}p{2cm}}
    \newcolumntype{m}{>{\centering\arraybackslash}p{1.1cm}}
    %----------------------------------------------------
    \begin{tabular}{lmmmm}
    \hline
    Parameter 
        & Stable-Reset 
        & Stable-No-Reset
        & UnStable-Reset
        & UnStable-NoReset \\
    \hline
    LR &$10^{-3}$&$10^{-2}$&$10^{-3}$& $10^{-3}$\\
    LR-SSM &$10^{-3}$&$10^{-2}$&$10^{-3}$& $10^{-3}$\\
    WD  &$10^{-3}$&$10^{-4}$&$10^{-4}$& $10^{-4}$\\
    \hline
    \end{tabular}
    %----------------------------------------------------
\end{table}

\subsubsection{Datasets}
\label{sec:appendix:datasets}

\paragraph{MSWC}
\label{sec:appendix:datasets:mswc}
For each class $500$ train, $100$ validation and $100$ test samples are used.     
Each audio sample is converted to signed ternary sequence i.e. a sequence of values $-1,0,1$ using the Speech2Spikes (S2S) \cite{speechtospikes} preprocessing algorithm with $20$ input channels and $201$ time steps \cite{yik2024neurobench}. The data is further processed by sum-binning with a non-overlapping moving window along its time dimension with bin-size of $4$.

\paragraph{DVS-Gesture}
\label{sec:appendix:datasets:dvs_gesture}
The dataset has $1342$ samples divided into $256$ test and $1086$ train samples.
To improve generalization, we follow EventSSM’s data augmentation transformations \cite{schöne2024scalableeventbyeventprocessingneuromorphic}, and use the convention of binning events into time-frames. 
We follow the standard convention of aggregating raw events into frames by binning them over fixed time windows. Specifically, we use the  framework of \cite{tonic} with a bin size of $\Delta = 20$ms, which aligns with values commonly used in the literature—for instance, \cite{subramoney2023efficientrecurrentarchitecturesactivity} use $\Delta =25$ms, and \cite{innocenti2021temporal} use up to $\Delta =20$ms. As a result, the samples of DVS-Gesture have a varying time-sequence length, an average of $340$ time steps. % $340 \pm 80$ 
For hyperparameter optimization, we used $10\%$ of the training set as a validation set.

\paragraph{sMNIST}
\label{sec:appendix:datasets:sMNIST}

MNIST consists of $60\,000$ training and $10\,000$ testing samples. Since sMNIST is a  sequential variant of MNIST, the same test and train split is used. For hyperparameter optimization, we used $10\%$ of the training set for validation. 

%\printbibliography
\let\url\nolinkurl
\bibliographystyle{IEEEtran}
\bibliography{references}
\end{document}